\documentclass[10pt,journal,compsoc]{IEEEtran}
\usepackage{graphicx}
\usepackage{pgfplots}
\usepackage{color}
\usepackage{hyperref}
\hypersetup{pdfinfo={
  Title=Tracking with multi-level features,
  Author={Roberto Henschel, Laura Leal-Taix\'{e}, Bodo Rosenhahn, Konrad Schindler},
  Subject=Multiple People Tracking,
  Keywords={Multiple People Tracking, Feature Integration, Quadratic cost function, Linear Programming, Spectral Clustering}}
}
\usepackage[]{units}

\usepackage{amsmath,amssymb} 
 \DeclareMathOperator*{\argmin}{arg\,min}
  \DeclareMathOperator*{\med}{med}

\ifCLASSOPTIONcompsoc
  \usepackage[nocompress]{cite}
\else
  \usepackage{cite}
\fi

\ifCLASSINFOpdf
\else
\fi

\begin{document}

\title{Tracking with multi-level features}

\author{Roberto~Henschel,
        Laura~Leal-Taix\'{e},
        	    Bodo~Rosenhahn,
        Konrad~Schindler
\IEEEcompsocitemizethanks{\IEEEcompsocthanksitem R. Henschel and B. Rosenhahn are with the TNT group at Leibniz University of Hannover, Germany.
\protect\\
E-mail: \{henschel,rosenhahn\}@tnt.uni-hannover.de
\IEEEcompsocthanksitem L.Leal-Taix\'{e} is with Computer Vision Group at Technical University Munich, Germany. \protect \\
E-mail: leal.taixe@tum.de
\IEEEcompsocthanksitem K.Schindler is with the Photogrammetry and Remote Sensing Group, ETH Z\"urich, Switzerland.  \protect \\
E-mail: konrad.schindler@geod.baug.ethz.ch
}
\thanks{Manuscript received July, 2016}}


\IEEEtitleabstractindextext{%
\begin{abstract}
We present a novel formulation of the multiple object tracking problem which integrates low and mid-level features. 
In particular, we formulate the tracking problem as a quadratic program coupling detections and dense point trajectories.
Due to the computational complexity of the initial QP, we propose an approximation by two auxiliary problems, a temporal and spatial association, where the temporal subproblem can be efficiently solved by a linear program and the spatial association by a clustering algorithm.
The objective function of the QP is used in order to find the optimal number of clusters, where each cluster ideally represents one person.
Evaluation is provided for multiple scenarios, showing the superiority of our method with respect to classic tracking-by-detection methods and also other methods that greedily integrate low-level features.
\end{abstract}

\begin{IEEEkeywords}
Multiple People Tracking, Feature Integration, Quadratic cost function, Linear Programming, Spectral Clustering
\end{IEEEkeywords}}

\maketitle
\IEEEdisplaynontitleabstractindextext

\IEEEpeerreviewmaketitle

\IEEEraisesectionheading{\section{Introduction}\label{sec:introduction}}

\IEEEPARstart{D}{etecting} and tracking people in a video is an important task of
computer vision, and its output is used in many applications such as 
autonomous driving, video surveillance or activity recognition.
A common approach to recover the trajectories of multiple people is
\emph{tracking-by-detection}: first a person detector is applied to
each individual frame to find the putative locations of people. Then,
these hypotheses are linked across frames to form trajectories.
By building on the advances in person detection over the last decade,
tracking-by-detection has been very successful \cite{lealcvpr2014,milantpami2014,zamireccv2012}.
But, at the same time, the dependence on detection results --
typically bounding boxes -- is also a main limitation.
State-of-the-art object detectors \cite{dollartpami2014,sadeghieccv2014,rennips2015} perform well in not too crowded environments, but
they still consistently fail in the presence of significant
occlusions.

Although ``connecting the dots'' supplied by a pedestrian detector is
convenient, a lot of potentially important information is lost along
the way.
In particular, the tracker does not use the actual image data, except sometimes in the form of rather weak appearance models to
discriminate different people.
%
Recently, a number of approaches have proposed to step away from the
standard paradigm \cite{milancvpr2015,fragkiadakieccv2012,chencvpr2014} and instead tackle the tracking problem more
directly, going straight from low-level image cues to trajectories.
Instead of using detections they base their processing on low- or
mid-level information such as dense point tracks \cite{broxeccv2010} or space-time supervoxels \cite{xueccv2012}, thus
also moving closer to the related problem of motion segmentation.

In this paper, we propose a principled \emph{global formulation 
  to integrate low- and
  mid-level features for multi-target tracking}.
We cast the coupled problems of \emph{(i)} temporally linking features into feature tracks, \emph{(ii)} grouping
them into individual moving targets (persons),
as one single
quadratic program (QP) with linear constraints.
%
%
Solving the QP directly is computationally and memory infeasible, therefore we approximate the problem by solving two subproblems: 
The temporal linking can be solved efficiently and optimally by linear
programming relaxation, whereas the grouping of features into individual persons
is found with spectral clustering. 
%
%
%
In this paper, we use detections as mid-level features and dense point tracklets (DPTs) as low-level features, since both are commonly used together in the literature,
and therefore we can directly compare with state-of-the-art methods.
Low-level features generally provide very accurate motion information, while mid-level features like detections provide the necessary 
structure information.    
%
%
%
%
To summarize, the \textbf{contribution} of the present paper is
three-fold:
\begin{itemize}
\item{
  We propose a global formulation for the integration of mid- and low-level features for multi target tracking.
    The problem is cast as a quadratic program (QP) with linear constraints, coupling
  the problems of (i) linking features into tracks and (ii) grouping
them into individual moving targets (persons).}
\item{Given that the initial QP is NP-hard and computationally too expensive to solve, we propose an approximation using a decomposition into an efficient linear program and
 a clustering step.}
  \item {We propose to use the QP objective function to robustly determine the number of clusters, which is always a delicate step in other approaches.}
\end{itemize}

\begin{figure}[t]
	\centering
	{\includegraphics[width=1\linewidth]{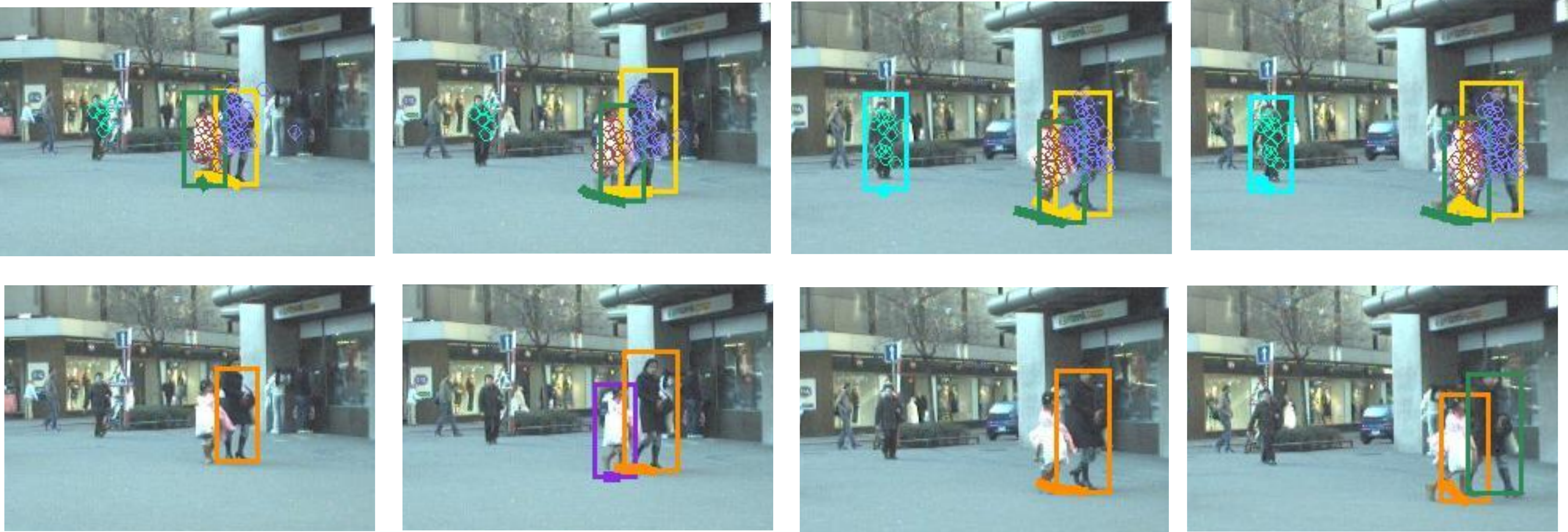}}
	\caption{Tracking results on ETH-Jelmoli over $18 $ frames. {\it Top row}: The proposed TbX tracker is able to generate stable trajectories. {\it Bottom row}: LP2D is unable to track both pedestrians consistently, creating ID switches and missed detections.}
	\label{fig:teaser}
\end{figure}

\subsection{Related work}

\noindent{\bf Tracking-by-detection.} Multiple people tracking is a
key problem for tasks such as surveillance or activity
recognition. Tracking-by-detection has become the standard way of
solving it. The problem is split into two steps: object detection and
data association.  In crowded environments, where occlusions are
common, even state-of-the-art detectors
\cite{dollartpami2014,galltpami2011,partbased} suffer from many false
alarms and/or missed detections.
The focus of data association is to overcome these detection failures,
by filtering out false alarms and
filling in the gaps due to false negatives.
While the data association step can be solved on a frame-by-frame
basis \cite{khantpami2005}, or one track at a time
\cite{berclazcvpr2006}, recent work has shown that it is beneficial to
jointly solve it over all tracks and all frames.
This is usually done in discrete space, using Linear Programming (LP)
\cite{jiangcvpr2007,zhangcvpr2008,pirsiavashcvpr2011,berclaztpami2011}
or graph-theoretic methods like generalized clique graphs
\cite{dehghancvpr2015} or maximum weight-independent sets
\cite{brendelcvpr2011}, although there are also continuous
formulations \cite{milantpami2014}.
Most of these methods can be solved optimally, though some not
efficiently, as they require the use of expensive decomposition methods
\cite{lealcvpr2012,buttcvpr2013}.  

All the aforementioned works are tracking-by-detection methods and
rely on a fixed set of detections as input. This has the drawback that
much of the image information is discarded during the non-maxima
suppression step built into any detector, potentially ignoring semi-occluded objects.
There have been some attempts to include additional evidence, by
starting from weaker candidate detections and coupling detection and
trajectory estimation \cite{leibeiccv2007}, using individual part
responses of DPM \cite{izadiniaeccv2012}, creating dedicated
detector for occluded people \cite{wojekcvpr2011}, or learning frequent
occlusion patterns \cite{tangiccv2013}.
Still, the basic problem remains, namely that the tracker is
completely dependent on the detector output, and has no access to the
potentially much richer image data.
We argue that tracking should be based on both mid-level features, such as a detector output, and low-level image information.\\

\noindent{\bf Tracking from low-level features.} 
In the recent literature, several works have started incorporating low-level image features for the task of multi-target tracking.
Few works use supervoxels as input for tracking, obtaining as a byproduct a silhouette of the pedestrian. In
\cite{chencvpr2014}, supervoxels are labelled according to target
identities with greedy propagation, starting from manually
initialised segmentation masks.
Greedy propagation tends to fail in crowded scenarios, leading
to long trajectories that often switch from person to person or from person to background.
Furthermore, the method needs manual initial segmentation masks. 
In
\cite{milancvpr2015}, supervoxel labeling is formulated as CRF
inference, where the targets are modelled as volumetric ``tubes''
through the sequence.
Here, we propose to assemble the tracking solution from the linking and clustering problem
which is much more efficient, and can integrate motion
information over much longer time windows than \cite{milancvpr2015}.
In \cite{chari2015pairwise}, the fusion of head and people detections to improve tracking performance is discussed. To this end, a quadratic program is used  to model non-maxima suppression as well as a simple overlap consistency between the different features. In contrast to our method, their model is designed to consider co-occurrences of active features only. Our formulation directly models the grouping of features to different persons, which is more appropriate for the tracking task, allowing to ensure consistency within each cluster. 
Note, that the input size of features and number of constraints in \cite{chari2015pairwise} is much smaller, allowing them to solve the relaxed version of the quadratic problem. Since we leverage DPTs, it is computationally not feasible to use their solver. Also in the extension \cite{Seguin16} to motion segmentation using superpixels, the per-person consistency is not considered. %
%
There are several works that use dense point tracks or KLT tracks together with detections to improve tracking performance.
Close to our work is \cite{benfoldcvpr2011}, where authors use the KLT feature tracker \cite{tomasiklt1991} as motion model to guide
  detection tracks frame-by-frame. The main difference to our method is that the treatment of both features is different, since each one has
  a specific purpose within their tracking framework. In contrast, we propose a holistic formulation, where both feature types can equally influence trajectory extraction. Let us consider the case where a bounding box is wrongly surrounding two people. 
  In \cite{benfoldcvpr2011}, the KLT features
would just provide conflicting information, but would not be able to generate two trajectories out of one bounding box
at that point in time. 
In contrast, the DPTs in time would be clustered into two different clusters, therefore creating two trajectories. 
%
Another related work is that of \cite{fragkiadakicvpr2011}, where multi-target
tracking is formulated as a clustering of dense feature tracks \cite{broxeccv2010}. This method suffers from two main problems: (i)
the automatic choice of the number of clusters on the basis of the eigengap
is unstable,
and (ii) individual tracks are typically very short on moving objects
and have a low temporal overlap among each other, which destabilises
the clustering and makes it impossible to recover from occlusion.
Follow up work was presented in \cite{fragkiadakieccv2012}, where dense point tracklets are combined
with detection-based tracklets in a two-step approach. 
In contrast, 
we propose a global optimization formulation to integrate low- and mid-level
features, so as to take full advantage of the strength of both.
The fact that we link the dense point tracklets across time to obtain longer,
but nevertheless reliable tracks, makes it possible for us to track even
through occlusions, unlike \cite{fragkiadakieccv2012}.
Recently, \cite{choiiccv2015} used interest point trajectories to create a more accurate affinity measure
to associate detections.


\section{Tracking with multi-level features}
\label{sec:tracking}

In this section, we present a formulation for the multi-target tracking problem which uses as input not only mid-level features such as detections, but also low-level features, e.g. dense point tracklets (DPTs) \cite{broxeccv2010}. 
Our goal is to define a common framework in which both features can be coupled to obtain a single set of trajectories that leverages the information of both feature channels.
Dense point tracks \cite{broxeccv2010} are capable of following a pedestrian even under partial occlusion, while detections give us the necessary structured information to distinguish pedestrians walking with a similar motion. Fig. \ref{fig:teaser}  demonstrates the advantage of this concept.

We formulate the multi-target tracking problem as a quadratic program (QP) where we couple: (i) temporal association of low- and mid-level features into {\it feature tracks}, (ii) spatial clustering of low-level features in accordance to the structure provided by the mid-level features (ideally one cluster per person).
Since such QP is NP-hard, we propose a decomposition that allows to find an approximate solution much more efficiently.

\begin{figure}[t]
\centering
{\includegraphics[width=1\linewidth]{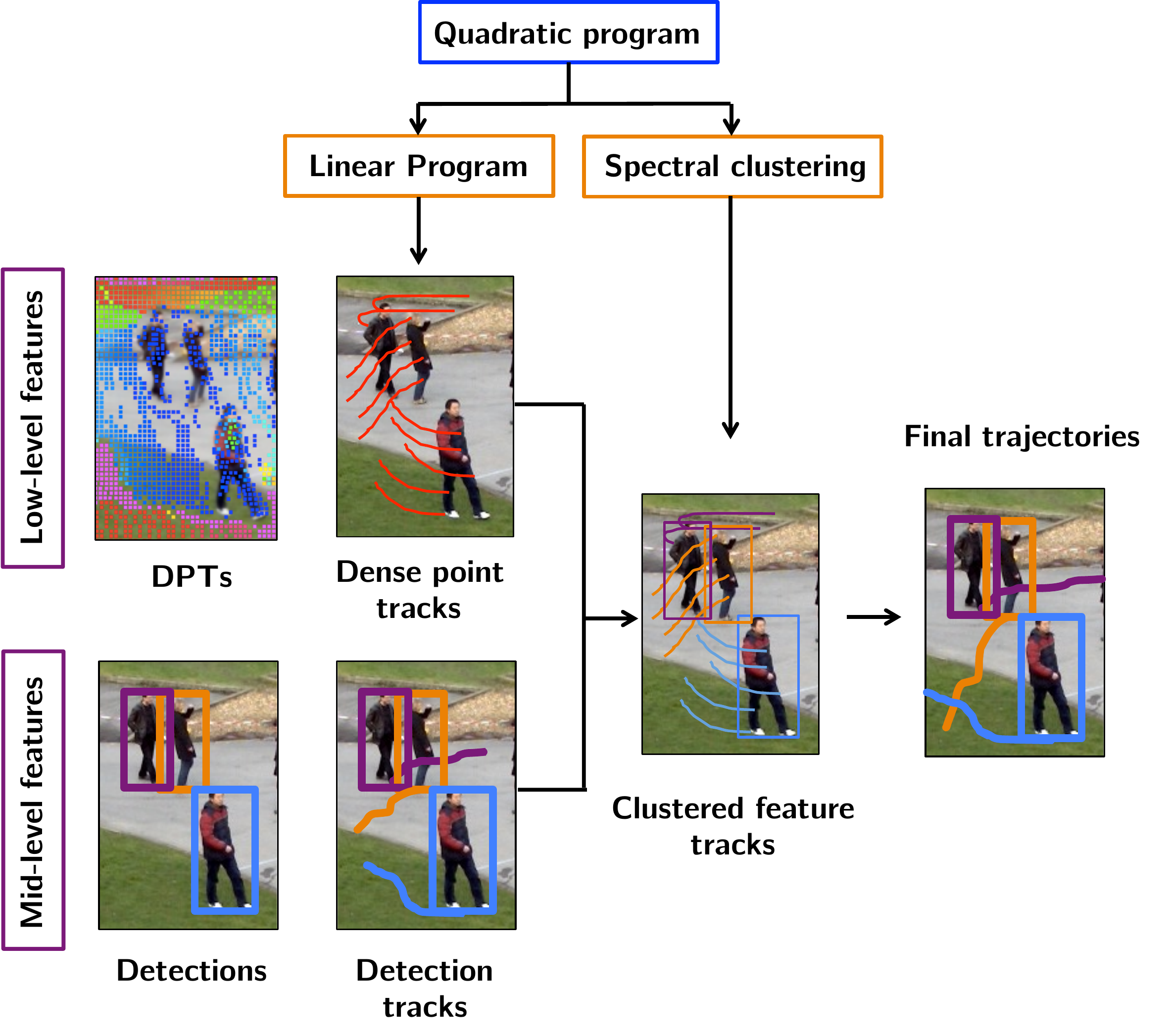}}
\caption{Diagram of our approach.}
\label{fig:diagram}
\end{figure}

\subsection{Temporal association with linear programming}
\label{linear}

We first focus on the temporal association of features into feature tracks. 
Let $\mathfrak{M}_{\mathfrak{c}}=\{\mathfrak{m}_i^{\mathfrak{c}}\}$ be a set of features of a category $c$. By way of example, we propose to use $\mathfrak{c} \in \mathfrak{C}=\{\text{low},\text{mid}\}$. For DPTs, i.e. $\mathfrak{c} = \text{low}$, 
$\mathfrak{m}_i^{\mathfrak{c}}=(\mathbf{p}^t_i,\mathbf{a}^t_i)$. At time $t=t_{i_j}$, $\mathbf{p}^t_i=(x,y)^T$ denotes its 2D image position, ${\bf a}^t_i$ the mean color value of a patch around $\mathbf{p}^t_i$, respectively. 
For $\mathfrak{c} = \text{mid}$, or detections in our case, features are defined as $\mathfrak{m}_j^{\mathfrak{c}}=(\mathbf{p}_j^t,w_j,h_j,t_j)$. At time $t=t_j$, detection $j$ is centered at $\mathbf{p}_j^t=(x,y)^T$ and has width $w_j$ and height $h_j$.
  
A feature track (of category $c$) is defined as an ordered list $T_n^{\mathfrak{c}} \subset \mathfrak{M}_c$ of features, sorted in time and without a temporal overlap.
Our goal is now to find the set of
tracks ${\mathcal{T}^{\mathfrak{c}}}*=\{T_n^{\mathfrak{c}}\}$ that best explain the feature evidence.

This can be formulated as a minimization of the following objective function:
\begin{align}
{\mathcal{T}^{\mathfrak{c}}}*&=\underset{ \mathbf{f}_\text{LP}^{\mathfrak{c}}}{\operatorname{{\bf argmin}}}  \quad \mathbf{c}_\text{LP}^{\mathfrak{c}} \mathbf{f}_\text{LP}^{\mathfrak{c}}\nonumber \\
&=\underset{ \mathbf{f}_\text{LP}^{\mathfrak{c}}}{\operatorname{{\bf argmin}}}  \sum_{i} c_\textrm{in}^{\mathfrak{c}} (i) f_\textrm{in}^{\mathfrak{c}} (i)  + \sum_{i} c_\textrm{out}^{\mathfrak{c}} (i)f_\textrm{out}^{\mathfrak{c}} (i) \nonumber \\
&+ \sum_{i} c_\textrm{d}^{\mathfrak{c}} (i) f_d^{\mathfrak{c}} (i)  +\sum_{i,j}  c_\textrm{t}^{\mathfrak{c}} (i,j) f_t^{\mathfrak{c}} (i,j) 
\label{eq:LP}
\end{align}
subject to edge capacity constraints, flow conservation at the nodes
\eqref{eq:lincon1}, \eqref{eq:lincon2} and exclusion constraints. See \cite{zhangcvpr2008} for further details.

The flags $ \mathbf{f_\textrm{LP}^{\mathfrak{c}}}$ take values in $\{0,1\}$, indicating whether a
particular feature connection is taken into the solution ($f=1$) or
not ($f=0$). 
The start/end costs $c_\textrm{in}^{\mathfrak{c}}$ and $c_\textrm{out}^{\mathfrak{c}}$ define how probable it is for a track to start
or end. These are learned from training data and kept the same for all the experiments in Section \ref{section:exps}.  
%
For detections, $c_\textrm{d}^{\text{mid}}$ will be proportional to the score given by the detector, so that only confident detections will be matched into tracks.
For DPTs, on the other hand, we constraint $f_\textrm{d}^{\text{low}}=1$ and set the costs $c_\textrm{d}^{\text{low}}=0$, so that they will all be matched.

The cost of a link edge $c_\textrm{t}^{\mathfrak{c}}(i,j)$ measures the affinity between features $\mathfrak{m}_i^\mathfrak{c},\mathfrak{m}_j^\mathfrak{c}$,
based on motion, appearance (in case of DPTs; for detections, using appearance is not beneficial) and temporal separation:
\begin{align}
c_\textrm{t}^{\mathfrak{c}}(i,j)  =  \frac{\| \mathbf{p}_j^{t+\Delta t} - \mathbf{p}_i^t \| }{V_\textrm{max}  \Delta t}  +  \frac{\| \mathbf{a}_j^{t+\Delta t} - \mathbf{a}_i^t \| }{A_\textrm{max} }  + \frac{ \Delta t  }{F_\textrm{max} }  
\label{costD}
\end{align}
where $V_\textrm{max}$ is the maximum speed of a pedestrian in pixels,
$A_\textrm{max}$ is the maximum appearance distance we allow, and
$F_\textrm{max}$ is the maximum time gap.
Eq.~\eqref{eq:LP} is the classic integer linear programming (ILP)
formulation, which has been extensively used in multiple object
tracking with detections as features. After relaxation to a linear program it can be efficiently
solved using Simplex \cite{zhangcvpr2008} or $k$-shortest paths
\cite{berclaztpami2011}.

\subsection{Quadratic terms for spatial association}
\label{quadratic}
For detections, the formulation of Eq.~\eqref{eq:LP} already provides suitable pedestrian trajectories. Nonetheless, since we also work with low-level features such as DPTs, a
straight-forward application of Eq.~\eqref{eq:LP} would yield many DPTs per pedestrian. 
We therefore need to impose further constraints, so as to obtain
larger clusters of coherent tracks which coincide with the detections.
Roughly speaking, we want to combine DPT tracks which are close to
each other and run in parallel, meaning they likely follow the same
person.
At the same time, we want to allow two DPT tracks to belong to
different clusters, if they only approach each other for a short period
of time, e.g., when two people cross.
Note, that it is in this step where the different features are coupled.
We encode these conditions in the form of a quadratic term in the
objective function, and an extra set of linear constraints. 

The goal now is to find the set of full trajectories $\mathcal{S}*=\{S_m\}$ which cluster tracks $T$ into pedestrians. 
This is expressed by 
the following quadratic problem:
\begin{subequations}
\label{eq:fullLP}
\begin{align}
&\mathcal{S}*=\underset{\bf f}{\operatorname{{\bf argmin}}}  \quad \mathbf{c}^\intercal \mathbf{f} +  \mathbf{f}^\intercal \mathbf{Q} \mathbf{f}, 
\\
&\text{subject to} \notag\\
&\qquad f_\textrm{in}^{\mathfrak{c}}(i) + \sum_{j} f_\textrm{t}^{\mathfrak{c}}(j,i) =  f_\textrm{d}^{\mathfrak{c}}(i) \, \forall \mathfrak{c} \in  \mathfrak{C}\label{eq:lincon1}\\
&\qquad f_\textrm{d}^{\mathfrak{c}}(i) =  f_\textrm{out}^{\mathfrak{c}}(i) + \sum_{j} f_\textrm{t}^{\mathfrak{c}}(i,j)  \, \forall \mathfrak{c} \in  \mathfrak{C} \label{eq:lincon2}\\
&\qquad \sum_{k=1 \ldots N_{cl}} f_\textrm{QP} (i,k) = f_\textrm{d}^{\mathfrak{c}}(i)  \, \forall \mathfrak{c} \in  \mathfrak{C} \label{eq:qpcon1}\\
&\qquad  f_\textrm{t}^{\mathfrak{c}} (i,j) + f_\textrm{QP} (i,k) -  f_\textrm{QP} (j,k)  \leq 1 \, \forall \mathfrak{c} \in  \mathfrak{C} \label{eq:qpcon2}\\
&\qquad f_\textrm{t}^{\mathfrak{c}} (i,j) - f_\textrm{QP} (i,k) +  f_\textrm{QP} (j,k)    \leq 1 \, \forall \mathfrak{c} \in  \mathfrak{C} . \label{eq:qpcon3}
\end{align}
\end{subequations}

The aforementioned constraints are created for all indices $i,j$ in the corresponding index sets, which we omitted for clarity.
Let $d_{\text{LP}}$ and  $d_{\text{QP}}$ define the number of  linear and quadratic costs, respectively.
The flag vector $\mathbf{f}$ is decomposed into the flags $\mathbf{f}_\text{LP}^{[\mathfrak{C}]} \in \{0,1\}^{d_{\text{LP}}}$ for the temporal association and  $\mathbf{f}_\text{QP} \in  \{0,1\}^{d_{\text{QP}}}$ for the spatial association, so $\mathbf{f} = \begin{pmatrix}
\mathbf{f}_\text{LP}^{[\mathfrak{C}]} & \mathbf{f}_\text{QP}
\end{pmatrix}^T$. Hereby we fixed an order $\{\mathfrak{c}_1,\cdots,\mathfrak{c}_{|\mathfrak{C}|}\}=\mathfrak{C}$ and set $\mathbf{f}^{[\mathfrak{C}]}:=(\mathbf{f}^{\mathfrak{c}_1},\cdots,\mathbf{f}^{\mathfrak{c}_{|\mathfrak{C}|}})$.

Using the costs defined in \eqref{eq:LP}, we set $\mathbf{c} = \begin{pmatrix}
\mathbf{c}_\textrm{LP}^{[\mathfrak{C}]} & \mathbf{0}
\end{pmatrix}^T$.
Now $\mathbf{Q}$ provides the costs for two features' spatial compatibility. We derive it from an affinity matrix $\mathbf{W}_\text{Spatial}$ that is described in detail in Sec.~\ref{sec:clustering}. Given $\mathbf{W}_\text{Spatial}$, we transform it into $\mathbf{Q}_\text{Spatial}:=-2\mathbf{W}_\text{Spatial}+\mathbf{1} $, so that $Q_\text{Spatial}(i,j) \in [-1,1]$, for all $i,j$, 
Finally, we construct the complete quadratic cost matrix via 
\begin{align*}
\mathbf{Q}&=\begin{pmatrix}
\mathbf{0}  & \mathbf{0} & 0 &  \mathbf{0}  \\
\mathbf{0}  & \mathbf{Q}_\text{Spatial} & 0 &  \mathbf{0}  \\
\mathbf{0}  & 0 & \ddots  & \mathbf{0}  \\
\mathbf{0}  & 0& \cdots& \mathbf{Q}_\text{Spatial} 
\end{pmatrix}.
\end{align*}
Note that $Q(i,j)=0$ for all $i,j  \leq d_\text{LP}$. Furthermore, we have $N_{cl}$ copies of $\mathbf{Q}_\text{Spatial}$ on the diagonal, $N_{cl}$ is an upper bound on the number of clusters.
Accordingly, we have $\mathbf{f}_\text{QP} \in  \{0,1\}^{N_\text{cl}F^2}$, where $F$ denotes the number of all feature tracks.

Now, for each $k\in \{0,\cdots ,N_\text{cl}-1\}$ and $i,j \in \{kF+1,\cdots,(k+1)F\} +d_\text{LP}$, the entry $Q(i,j)$ is the cost of assigning both nodes $i$ and $j$ to cluster
$k$. \\

\noindent{\bf Coupling LP and QP terms through the constraints.} 
The constraint \eqref{eq:qpcon1} guarantees that each active tracklet  $i$  (i.e., $f_\textrm{d}^{\mathfrak{c}}(i)=1$) is
assigned to exactly one cluster. 
Whenever tracklet $i$ and $j$ both appear in the same cluster $k$, the
variables $f_\textrm{QP}(i,k)$ and $f_\textrm{QP}(j,k)$ will both be
active, and the cost $Q(i,j)$ will be applied. If only one of the tracklets is assigned to the cluster, e.g. $f_\textrm{QP}(i,k)=0$ and $f_\textrm{QP}(j,k)=1$,
the resulting cost will be zero.
This definition of costs will drive the optimization towards a
solution which is temporally consistent according to the temporal costs in $\mathbf{c}^\intercal \mathbf{f}$, and at the same
time will try to cluster similar tracklets, which are likely to belong to the same moving object.

Eqs. \eqref{eq:qpcon2} and \eqref{eq:qpcon3}, on the other hand, guarantee the temporal
consistency of the clusters. That is, if tracklets $i$ and $j$ are
linked in time by edge $f_\textrm{t}^{\mathfrak{c}}(i,j)$, then both tracklets belong
to the same cluster $k$.
Note that the linear and quadratic terms are coupled through the three constraints \eqref{eq:qpcon1}, \eqref{eq:qpcon2}, and \eqref{eq:qpcon3}. \\
 
\noindent{\bf Final trajectories.} 
Finally we create a pedestrian trajectory $S_m$ for each cluster if it contains features from both categories (detections and DPTs). DPTs on the 
background are grouped into outlier clusters (since no detections are associated with them) which are then removed. Details are described in Sec. \ref{subsec:trajectory_extraction}.
Conceptually, DPTs help to distinguish multiple people (with different motion) inside the same bounding box and detections help in distinguishing people who have the same motion pattern but are inside different detections. Due to the constraints (3d)-(3f) consistency between the feature categories is ensured.

\subsection{Approximation of the problem}

The program defined in Eq. \ref{eq:fullLP} has one practical drawback: as a quadratic
program it is NP-hard. We can solve it with branch-and-bound, but this
is computationally inefficient and unfeasible even if we process a sequence in small batches. 
Alternatively, one can convert each quadratic constraint into three
linear constraints \cite{torresanieccv2008}, bringing the program
to linear form. However, the corresponding constraint matrix is
not totally unimodular (unlike the constraints of the LP \eqref{eq:LP}), which
means that its LP-relaxation is not tight, and one is again faced with
a NP-hard ILP.
One can still relax the integrality constraint, solve the problem, and later apply 
a rounding scheme as in \cite{charicvpr2015}. Nonetheless, even in that case,
the initial linearized problem has millions of constraints even if we track only few pedestrians in few frames,
which means finding a solution is computationally infeasible.
Finally, one could resort to decomposition methods like dual
decomposition \cite{torresanieccv2008} or Dantzig-Wolfe
\cite{lealcvpr2012}, but overall the
problem remains unfeasible memory and computation-wise. \\

\noindent{\bf Proposed solution.} We propose to take advantage of the particular structure of
our program and divide the optimization into two steps as shown in Fig. \ref{fig:diagram}. In the first
step, we solve the initial LP of Eq. \eqref{eq:LP}, which will
result in a set of low- and mid-level feature tracks $\mathcal{T}^{\mathfrak{c}}=\{{T}_n^{\mathfrak{c}}\}$.

In a second step, we approximate the minimization of the quadratic part of the cost function \eqref{eq:fullLP},
given the linked tracklets by the linear solution. This part can be considered as a correlation clustering problem \cite{bansal2004correlation}, which we tried to solve directly. 
However, the solvers did not terminate even after days of computation, even though applicability on a large scale is suggested \cite{bagon2011large}.

Since correlation clustering seems to be too expensive for our task, we use a sample-based approach. We generate clusters using spectral clustering \cite{shi2000normalized,cour2004normalized}, which is known to provide good quality results, and evaluate it using the quadratic cost function defined by $\mathbf{Q}$. 
%

A key aspect of most clustering methods is the
choice of the number of clusters $k$, which varies from one problem
instance to the next, and thus needs to be determined in a data-driven
fashion.
We propose to choose the best $k$ by computing the set of trajectories
for several values of $k$, and selecting the best of these candidate
sets, based on the cost \eqref{eq:fullLP} of the full quadratic model.
The function \eqref{eq:fullLP} describes our complete tracking
problem, and is thus more suitable as a quality metric than more
heuristic measures for the goodness of a given clustering.
In fact, the proposed strategy bears some similarity with
``re-ranking'' strategies in the field of recognition and detection,
which also replace a sophisticated, but computationally costly model
with a simpler proxy to obtain a list of candidate solutions, and then
rescore those candidates with the full model.
In our case decomposing the problem has a further advantage: once feature tracks
 with high temporal overlap have been found, one can
easily extract rich longer-term motion descriptors and consequently
obtain a better clustering than with only short feature tracklets
\cite{broxeccv2010} or single detections.
Using that information is intractable in the integrated model
\eqref{eq:fullLP}, since it would introduce a huge (combinatorial)
number of higher-order terms.

\section{Clustering}
\label{sec:clustering}

Using the temporal associations from the decomposed linear program, the tracking formulation \eqref{eq:fullLP} simplifies to solving the quadratic costs given the constraints \eqref{eq:qpcon1}-\eqref{eq:qpcon3}. This however can be seen as a clustering problem that is well-approximated by standard clustering algorithms, e.g., normalized cuts
\cite{shi2000normalized,cour2004normalized}, where the number of clusters $k$ has to be known a priori.
The current state-of-the-art tracking solution \cite{fragkiadakieccv2012} that follows a similar clustering approach, is lacking of a good strategy to compute this parameter. A heuristic approach is used, namely several normalized cuts are computed with varying parameters and combine various clusters from different n-cut results to construct the final clusters.

In contrast, we follow a systematic approach with a reduced search space compared to \cite{fragkiadakieccv2012}.
We propose a computationally reasonable approach to find a proper $k$, that is derived directly from our problem formulation \eqref{eq:fullLP}, thus providing a rational parameter inference. Moreover, the number of detection tracklets can be used as a good initial guess for the parameter $k$.
Ideally, each cluster will correspond to a trajectory, composed of several low-level features which will group into the structure provided by the mid-level features.
In the following, we discuss
our solution in more detail.

\subsection{Defining the distance matrix}

Between any two features tracks $T_i$ and $T_j$, we compute an affinity $W_\text{Spatial}(i,j) \in [0,1]$ according to their motion and spatial consensus.
%
While \cite{fragkiadakieccv2012} takes detection tracks as tracking units, we keep their temporal association as a soft constraint, thereby avoiding error propagation from the feature tracks obtained by the LP formulation, i.e. the clustering can implicitly compensate from temporal linking errors.

There are three different types of affinities that we consider: $\mathbf{W}_\text{PP}$ between dense point tracks (detection overlap, distance, speed and angle), $\mathbf{W}_\text{PD}$ between dense point tracks and detection tracks (intersection, distance, speed, angle) and $\mathbf{W}_\text{DD}$ between detections tracks (intersection and DPT overlap). Note that in consistence with the temporal linking, we use only those detections that have not already been removed by the LP in Eq. \eqref{eq:LP}.
The entire affinity matrix is then given by
\begin{align}
\mathbf{W}_\text{Spatial}=\begin{pmatrix} \mathbf{W}_\text{PP} & \mathbf{W}_\text{PD} \\ 
\mathbf{W}_\text{PD}^T & \mathbf{W}_\text{DD}
\end{pmatrix}.
\end{align}
It is worth mentioning that $\mathbf{W}_\text{Spatial}$ can be easily expanded to include more feature categories in a similar manner.
Special care has to be taken to define values of $\mathbf{W}_\text{Spatial}$ when the information in the image data is insufficient to define a reasonable affinity value. If we would simply set the affinity to $0$, 
it would mean that the affinity in the absence of any
information is lower than if the two trajectories have strongly
incoherent clues. 
Our compromise for such cases is to assign the value $0.5$, with the idea that it should not have a tendency for either being clustered or separated in different clusters.
The details of the affinity definition can be found in the Appendix.

\subsection{Cluster evaluation}

Once we have a properly set affinity matrix, we can apply normalized cuts to obtain the spatial links.
The last question that remains is how to set the number of clusters
$k$ correctly.
There are heuristics to automatically estimate $k$, such as
\cite{li2007noise,zelnik2004self}, but these seem to be somewhat
problem-specific and did not work well in our case.
Instead, we propose to go back to the original quadratic program
\eqref{eq:fullLP}, and use the actual objective of our tracking task
to determine the cluster number.
The affinity matrix $\mathbf{W}_\text{Spatial}$ is transformed into costs via $\mathbf{Q}_\text{Spatial}=-2\mathbf{W}_\text{Spatial}+\mathbf{1}$ so that it has costs in $[-1,1]$.   As minimizing the quadratic costs defined by $\mathbf{Q}_\text{Spatial}$ results in a correlation clustering problem that is hard to solve, we generate cluster samples using spectral clustering in order to minimize the cost function.
In particular this means we compute clustering results for different $\hat{k}$ using spectral clustering. Each result with cluster number $\hat{k}$ corresponds to a decision vector $\mathbf{f}_{\hat{k}}$.
The optimal number $k$ together with the cluster is then found according to the original objective function via 
\begin{align}
k:=\argmin_{\hat{k}=1,\ldots,N_{\text{cl}}}{\mathbf{f}_{\hat{k}}^T \mathbf{Q}_{\text{Spatial}} \mathbf{f}_{\hat{k}}}.
\end{align}

\subsection{Trajectory extraction}
\label{subsec:trajectory_extraction}
In order to generate the final trajectories, we first connect the clustered detections. We use the clustered dense point tracks to obtain reliable interpolations between detections and extrapolations, as long as we have reliable dense point tracks in a cluster, indicating the same direction, i.e., having a low variance in their direction. Furthermore, clusters which do not contain detections are considered as outlier clusters, i.e., background DPTs, and are thus ignored.

	\section{Experimental results}
\label{section:exps}

We evaluate the performance of the proposed algorithm in three parts: (i) the correctness of the dense point tracklets obtained with LP, (ii) evaluation of the different parts of our method, and (iii) the tracking performance in several publicly available datasets, as well as the challenging MOTChallenge benchmark \cite{motchallenge:arxiv:2015} which contains 11 sequences for testing.  \\

\noindent{\bf Implementation details.} We process the videos in batches of 50 frames. For each batch, we compute the n-cut result using \cite{shi2000normalized,cour2004normalized}. Since n-cut involves a random process, we run the clustering for each cluster parameter $50$ times and take the best result, using our cluster evaluation function. 
All experiments are performed with: $\sigma_\text{Dist}=0.4,\mu_\text{Dist}=0.05$ and $\sigma_\text{angle}=50$.  For performance reasons, we filter out a DPT if it does not indicate any motion or if the response of the detection's confidence map indicates that it lies on the background. We used $V_\textrm{max} = \unitfrac[25]{pixel}{s}, F_\textrm{max}=15$ and $A_\textrm{max} = 20$.\\

\noindent{\bf Performance evaluation.}
For the subsequent tracking performance evaluation, we use the popular CLEAR MOT metrics \cite{clear} that provide two complementary measures: tracking accuracy (TA), which incorporates missing recall, false alarms and identity switches; and tracking precision (TP) which is a measure of the localization error (overlap of bounding boxes if the evaluation is done in 2D or distance between detections). The overlap to consider a detection a match in the ground truth is 50\%. 
We also quote four popular
metrics proposed in \cite{licvpr2009}. The first two reflect the
temporal coverage of true trajectories by the tracker: mostly tracked (MT, $> 80\%$ overlap with ground truth) and mostly lost (ML,
$< 20\%$), while the third and fourth are the well-known Recall and Precision measures.\\

\subsection{Dense point tracks}
\label{exps1}

Firstly, we evaluate the correctness of the dense point tracklets obtained by solving the LP and compare them to the initial DPTs of \cite{broxeccv2010}. 
For this, we use the Figment dataset \cite{fragkiadakicvpr2011}, which contains 18 sequences of basketball players and ground truth masks every 7 frames. 
Making this experiment on a sequence with segmentation masks instead of bounding boxes allows us to make sure that we are not counting as moving object the part of the background that is inside the bounding box.

We present three measures: (i) IDsw/traj: that is how many tracks switch from background to person (or viceversa) or between two different players, averaged per track; (ii) Avg. length: average length of a track, measured as a percentage of the total length of the sequence; (iii) Overlap: average number of frames of overlap between tracks.

As we can see in Table \ref{tab:longtraj}, the tracks that we generate are much longer than the ones from \cite{broxeccv2010}, with an average length of almost 80\% of the sequence. This comes at the expense of a few identity switches (or leaking as referred to in \cite{fragkiadakieccv2012}), showing that the proposed tracks are robust enough for clustering. 
More importantly, the overlap between tracks is increased from $0.7$ to $5.16$ frames, meaning the affinities computed from these tracks are much more meaningful. 

\begin{table}[tbh]
    \caption{Comparison of feature tracks vs DPTs.}
    \begin{center}
     \begin{tabular}{l | c c c}
      Method & IDsw/traj & Avg. length  & {Overlap}  \\ \hline
         \cite{broxeccv2010}  &  $\mathbf{8.5 \times 10 ^{-4}}$ & 30.15 & 0.74   \\
    Proposed   &  $1.0 \times 10 ^{-1}$ & {\bf 79.94} & {\bf 5.16} \\  \hline  
    \end{tabular}
    \end{center}
\label{tab:longtraj}
\end{table}

\subsection{Analysis of the proposed method}
\label{exps3}
In these experiments, we analyze the performance of each of the components of the proposed method. We first perform experiments with the $k$ obtained from the QP objective function, and analyze the contribution of each of the affinities described in the Appendix. 
We consider several baselines that consists of the following parts:
\begin{itemize}
\item LP2D: Using only the standard LP-approach \eqref{eq:LP} on detections.
\item Dist: Using detections and DPTs. As affinities between DPT's the spatial distance (Eq. \ref{eq:Dist}) is considered.
\item Det: Using detections and DPTs. As affinities between DPT's  the box-driven distance (Eq. \ref{eq:Det}) is considered.
\item Speed: Using detections and DPTs. As affinities between DPT's the speed affinity (Eq. \ref{eq:Speed}) is considered.
\item Angle: Using detections and DPTs. As affinities between DPT's the angle affinity (Eq. \ref{eq:Angle}) is considered.
\end{itemize}

Furthermore, we compare to \cite{fragkiadakieccv2012} (Greedy), which also incorporates detections and dense points tracks, but computes the number of clusters in a greedy fashion.
As datasets we use 6 sequences: TUD-Stadtmitte, PETS09-S2L1, S2L2, S2L3, S1L1-2, S1L2-1 and report the average performance over the 6 sequences given as MOTA score in Table \ref{tab:baselines}. 
As we can see, our search approach for the optimal number of clusters clearly outperforms the greedy one, by more than 15\%. Furthermore our coupling formulation of low- and mid-level features can successfully integrate information from both channels, and thus improves over a tracking system that is based on detections only (LP2D) by more than 3\%.
Finally, Table \ref{tab:baselines} shows that clustering feature tracks leverages from the combination of all defined affinities, i.e. we can successfully group DPTs belonging to the same person by taking distance and motion information into consideration.\\
 
\begin{table}[b]
    \caption{Comparison to baseline}
    \begin{center}
     \begin{tabular}{l | c c c}
      Affinity & MOTA \\ \hline
      Greedy (detections+DPT) \cite{fragkiadakieccv2012} & 22.3 \\ 
      LP2D (detections) & 34.8 \\ 
      Dist & 35.8 \\
      Det & 36.5 \\
      Dist+Det & 36.5 \\ 
      Dist+Det+Speed  & 36.3 \\
      Dist+Det+Angle  & 36.6 \\
      Dist+Det+Angle+Speed & {\bf 37.9 } \\ \hline
    \end{tabular}
    \end{center}
\label{tab:baselines}
\end{table}

\begin{figure}[b]
\begin{center}
\begin{tikzpicture}[scale=0.7]
\begin{axis}[
  xlabel=Number of clusters,
  ylabel=MOTA,
  only marks,
  axis background/.style={fill=gray!10},
  x=0.9cm,
  y=0.2cm,
  xtick=data,
xticklabels={$k-5$,$k-4$,$k-3$,$k-2$,$k-1$,$k$,$k+1$,$k+2$,$k+3$,$k+4$,$k+5$},
]
\addplot table {kVariation.dat};
\end{axis}
\end{tikzpicture} 
\end{center}
\caption{Variation around the computed number of clusters on the 6 sequence dataset.}
\label{fig:variation}
\end{figure}
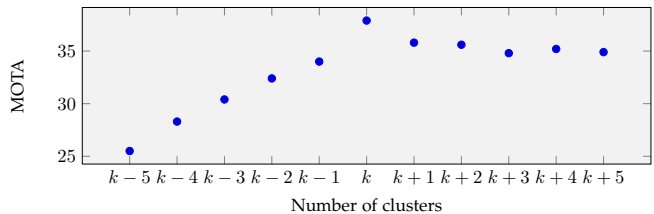

\noindent{\bf Choosing the best $k$ for spectral clustering. }
We also evaluated if we can correctly find the best number of clusters with our QP formulation. To this end, we vary the number $k'$ of clusters around the number $k$, chosen by our method
and compute the averaged MOTA score on the 6 sequences. As shown in Fig. \ref{fig:variation}, our method automatically selects the number of clusters that provides the best tracking accuracy.

\subsection{Multiple people tracking}
\label{exps2}

Finally, we evaluate the proposed method against state-of-the-art trackers on 35 sequences. 
For all experiments, we use only publicly available detections and ground truth, and use the evaluation scripts from \cite{motchallenge:arxiv:2015}. 

We compare to several state-of-the-art methods, drawing special attention to:
\begin{itemize}
\vspace{-0.1cm}
\item Comparison with \cite{fragkiadakieccv2012,benfoldcvpr2011}: both methods use detections as well as low-level trajectories (DPTs and KLT feature tracks, respectively). 
\item Comparison to LP2D: trajectories obtained by using only the Linear Program part of our algorithm, which links detections using image (2D) information.
\end{itemize}

%
%
The first set of sequences of our experiments is the Urban dataset \cite{fragkiadakieccv2012}. This contains 17 urban crowded scenes filmed from a pedestrian's viewpoint, which creates complex occlusions and pedestrians of various sizes cross the field of view of the camera. 
This is why standard detectors \cite{dollarpami2014} struggle to find the pedestrians in these sequences, and why accuracy is in general low. We convert the ground truth of  \cite{fragkiadakieccv2012} from segmentation masks to bounding boxes, so that we can compare with the aforementioned metrics.
The results are shown in Table \ref{tab:urban}. Note, that even though both methods presented use low- and mid-level features, we outperform the method in all measures, specially reducing the identity switches from 173 to 28. 

\begin{table}[tbh]
    \caption{Results on URBAN dataset. }
    \begin{center}    
\begin{tabular}{l|c c c c}
  Method & TA & TP & IDsw & Frag \\ \hline
    Fragkiadaki et al. \cite{fragkiadakieccv2012} &  5.0 & 73.2  & 173 & 180  \\ 
    Proposed   &  {\bf 9.1} & {\bf 74.2} &  {\bf 28} & {\bf 41}\\  \hline  
    \end{tabular}  
    \end{center}
\label{tab:urban}
\end{table}

Next we test 6 sequences: TUD-Stadtmitte, PETS09-S2L1, S2L2, S2L3, S1L1-2, S1L2-1. As shown in Table \ref{tab:6seqs}, top part, we outperform all methods in tracking accuracy (TA). We track significantly more trajectories and have less identity switches than competing methods. 
Note that \cite{fragkiadakieccv2012}, that also uses low-level features, has less ID switches, but its tracking accuracy is 15\% lower than our method and it is able to track less than half of our trajectories (MT), proving that our proposed formulation is much superior at combining low- and mid-level features.
We also compare our method on AVG-TownCentre \cite{benfoldcvpr2011}, a sequence of a busy city center filmed from a high viewpoint, with detections from \cite{benfoldcvpr2011}.
As we can see in Table \ref{tab:6seqs}, we outperform \cite{benfoldcvpr2011}, a method that uses low-level features in the form of KLT tracks, by 6 percentage points in accuracy. The precision of the method is higher for \cite{benfoldcvpr2011} because they do bounding box position refinement, while we directly use the bounding boxes output by the detector.
We also improve over tracking-by-detection methods like LP2D in accuracy, precision and identity switches. 
%

\begin{table}[tb]
\caption{Evaluation in image space. {\it Top:} Averaged over six sequences: 
PETS09-S2L1, S2L2, S2L3, S1L1-2, S1L2-1, TUD-Stadtmitte. {\it Bottom:} Results on AVG-TownCentre.}
\begin{center}
\begin{tabular}{l|p{0.4cm} p{0.4cm}p{0.4cm}p{0.4cm}p{0.4cm}p{0.4cm}p{0.4cm}}
\hline
 \multicolumn{8} {c} { 6 sequences} 
  \\ \hline
  Method & TA & TP & Rcll & Prcn & MT & ML & IDsw \\ \hline

  Fragkiadaki et al. \cite{fragkiadakieccv2012} & 22.3 & {\bf 72.1} & 35.5 & 74.1 & 8.4 & 44.1 & {\bf 216}  \\  
  Pellegrini et al. \cite{pellegriniiccv2009} & 28.5 & 65.2 & 48.6 & 75.1 & 15.8 & 31.7 & 1162  \\  
 Yamaguchi et al. \cite{yamaguchicvpr2011} & 29.0 & 65.1 & 48.8 & 74.8 & 16.3 & {\bf 31.2} & 987  \\  
  Leal-Taix{\'e} et al. \cite{lealiccv2011} & 31.9 & 65.7 & 50.4 & 75.5 & 19.8 & 34.2 & 608  \\  
 LP+2D & 34.8 & 65.4 & 50.0 & 79.5 & 18.3 & {\bf 31.2} & 663  \\  
    Proposed & {\bf 37.9}  & 65.5 & {\bf 50.9} & {\bf 81.4} & {\bf 20.3} &  32.2 & 414 \\ \hline 
    \hline
    
     \multicolumn{8} {c} { AVG-TownCentre} 
  \\ \hline
  Method & TA & TP & Rcll & Prcn & MT & ML & IDsw \\ \hline
   Yamaguchi et al. \cite{yamaguchicvpr2011} & 52.8 & 76.6 & 77.3 & 79.5 & 50.4 & 7.5 & 328  \\  
    Pellegrini et al. \cite{pellegriniiccv2009} & 55.2 & 76.6 & 78.6 & 80.7 & 54.0 & {\bf 7.1} & 324  \\  
          Benfold et al. \cite{benfoldcvpr2011} & 58.6 & {\bf 83.6} & {\bf 79.0} & 82.2 & {\bf 59.7} & 10.6 & 236  \\  
    LP2D & 61.6 & 77.2 & 73.9 & {\bf 89.3} & 48.7 & 12.4 & 245  \\  
    Proposed & {\bf 64.2}  & 79.1 & 76.9 & {\bf 89.3} &  51.8 &  9.3 & {\bf 231} \\
\hline
\end{tabular}
\end{center}
\label{tab:6seqs}
\end{table}

We also test on the recent MOTChallenge benchmark \cite{motchallenge:arxiv:2015}, which contains 11 sequences for training and 11 for testing. Sequences vary in viewpoint, density of pedestrians as well as moving/static cameras, which makes it extremely hard for trackers to work well in all scenarios. 
For many of these sequences, pedestrians are very small, a very hard scenario for our tracker as there are very few dense tracks per pedestrian. 
In Table \ref{tab:motcha}, we provide the results as shown on the website. We are one of the best performing published algorithms in terms of tracking accuracy. Note, that unlike competing methods, we do not use any type of trajectory post-processing. 
We want to especially point out that we outperform LP2D by almost 8 percentage points in accuracy and less than half the identity switches, clearly showing the strength of combining low- and mid-level features.

\begin{table}[tb]
    \caption{Results on the MOTChallenge test set.}
\begin{center}
 \begin{tabular}{l|  c c c c c c  }
    Method   & TA & TP  & MT & ML  &  IDsw & FP \\ \hline
    NOMT \cite{choiiccv2015} & 33.7 & 71.9 & 12.2 & 44.0 & 442 & 7762 \\
    TDAM \cite{yang2016temporal} & 33.0 & 72.8 & 13.3 & 39.1 & 464 & 10064 \\
    MHT-DAM \cite{kimiccv2015}& 32.4 & 71.8 &  16.0 & 43.8 & 435 & 9064 \\
    MDP \cite{xiangiccv2015} & 30.3 & 71.3 & 13.0 &  38.4 & 680 & 9717 \\ \hline
        TbX (proposed)  & 27.5 &  70.6 & 10.4 & 45.8 & 759 & 7968 \\ \hline
    LP-SSVM \cite{wangbmvc2015}  & 25.2 & 71.7 & 5.8 & 53.0 & 849 & 8369 \\
    ELP \cite{mclaughlinwacv2015}& 25.0 & 71.2 & 7.5 & 43.8 & 1396 & 7345 \\
    JPDA-m \cite{rezatofighiiccv2015} & 23.8 & 68.2 & 5.0 & 58.1 & 365 & 6373 \\ 
    MotiCon \cite{lealcvpr2014} & 23.1 & 70.9 & 4.7 & 52.0 & 1018 & 10404 \\
    SegTrack \cite{milancvpr2015} & 22.5 & 71.7 & 5.8 & 63.9 & 697 & 7890 \\ 
    LP2D (baseline)  & 19.8 & 71.2 & 6.7 & 41.2 & 1649 & 11580  \\
    DCO-X \cite{milantpami2016} & 19.6 & 71.4 & 5.1 & 54.9 & 521 & 10652 \\
    CEM \cite{milantpami2014} & 19.3 & 70.7 & 8.5 & 46.5 & 813 & 14180 \\
	RMOT \cite{yoonwacv2015} & 18.6 & 69.6 & 5.3 & 53.3 & 684 & 12473 \\
	SMOT \cite{dicleiccv2013} & 18.2 & 71.2 & 2.8 & 54.8 & 1148 & 8780 \\
	ALExTRAC \cite{bewleyicra2016} & 17.0 & 71.2 & 3.9 & 52.4 & 1859 & 9233 \\
	TBD \cite{geigerpami2014} & 15.9 & 70.9 & 6.4 & 47.9 & 1939 & 14943 \\
	TC-ODAL \cite{baecvpr2014} & 15.1 & 70.5 & 3.2 & 55.8 & 637 & 12970 \\
	DP-NMS \cite{pirsiavashcvpr2011} & 14.5 & 70.8 & 6.0 & 40.8 & 4537 & 13171 \\
	LDCT \cite{soleraiccv2015} & 4.7 & 71.7 & 11.4 & 32.5 & 12348 & 14066 \\ \hline
    \end{tabular}
  \end{center}

\label{tab:motcha}
\end{table}

\section{Conclusion}
We presented a global formulation for the integration of mid- and low-level features for multi-target tracking. The problem is cast into a quadratic program, which is then decomposed for efficiency into temporal associations with linear programming and spatial associations with spectral clustering. The full objective function of the QP is still used to choose the optimal number of clusters for spectral clustering, which is always a delicate step in other approaches.
We showed superior results when compared to tracking-by-detection methods and methods that also use low-level features, thus proving the benefits of our formulation.
As future work, we plan on exploring the integration of other features. Since the number of affinities grows quadratically with the number of features, we will focus on metric learning for the affinity computation.


%





\ifCLASSOPTIONcaptionsoff
  \newpage
\fi

\bibliographystyle{IEEEtran}
\bibliography{tpami2016,refs-rob}



@Article{cour2004normalized,
  Title                    = {Normalized cut segmentation code. Copyright 2004 University of Pennsylvania},
  Author                   = {Cour, Timothee and Yu, Stella and Shi, Jianbo},
  Journal                  = {Computer and Information Science Department},
  Year                     = {2004}
}

@InProceedings{li2007noise,
  Title                    = {Noise robust spectral clustering},
  Author                   = {Li, Zhenguo and Liu, Jianzhuang and Chen, Shifeng and Tang, Xiaoou},
  Booktitle                = {ICCV},
  Year                     = {2007},
  Organization             = {IEEE},
  Pages                    = {1--8}
}

@Article{shi2000normalized,
  Title                    = {Normalized cuts and image segmentation},
  Author                   = {Shi, Jianbo and Malik, Jitendra},
  Journal                  = {TPAMI},
  Year                     = {2000},
  Number                   = {8},
  Pages                    = {888--905},
  Volume                   = {22},

  Publisher                = {IEEE}
}

@InProceedings{zelnik2004self,
  Title                    = {Self-tuning spectral clustering},
  Author                   = {Zelnik-Manor, Lihi and Perona, Pietro},
  Booktitle                = {Advances in neural information processing systems},
  Year                     = {2004},
  Pages                    = {1601--1608}
}

@article{bagon2011large,
  title={Large scale correlation clustering optimization},
  author={Bagon, Shai and Galun, Meirav},
  journal={arXiv preprint arXiv:1112.2903},
  year={2011}
}
@article{bansal2004correlation,
  title={Correlation clustering},
  author={Bansal, Nikhil and Blum, Avrim and Chawla, Shuchi},
  journal={Machine Learning},
  volume={56},
  number={1-3},
  pages={89--113},
  year={2004},
  publisher={Springer}
}
@article{yang2016temporal,
  title={Temporal Dynamic Appearance Modeling for Online Multi-Person Tracking},
  author={Yang, Min and Jia, Yunde},
  journal={Computer Vision and Image Understanding},
  year={2016},
  publisher={Elsevier}
}
@inproceedings{chari2015pairwise,
  title={On pairwise costs for network flow multi-object tracking},
  author={Chari, Visesh and Lacoste-Julien, Simon and Laptev, Ivan and Sivic, Josef},
  booktitle={Proceedings of the IEEE Conference on Computer Vision and Pattern Recognition},
  pages={5537--5545},
  year={2015}
}
@InProceedings{Seguin16,
    author = "Seguin, Guillaume and Bojanowski, Piotr and Lajugie, Rémi and Laptev, Ivan",
    title = "Instance-level video segmentation from object tracks",
    booktitle= "Proc. CVPR",
    year = "2016"
}









@article{rennips2015,
	Author = {S. Ren and K. He, R. Girshick and J. Sun},
	Date-Added = {2016-07-24 16:25:14 +0000},
	Date-Modified = {2016-07-24 16:25:14 +0000},
	Journal = {Neural Information Processing Systems (NIPS)},
	Owner = {lealtaix},
	Timestamp = {2016.07.18},
	Title = {Faster R-CNN: Towards Real-Time Object Detection with Region Proposal Networks},
	Year = {2015}}

@book{networkflows,
	Author = {R.K. Ahuja and T.L. Magnanti and J.B. Orlin},
	Date-Added = {2011-11-20 12:51:34 +0100},
	Date-Modified = {2011-11-20 12:52:25 +0100},
	Owner = {laurale},
	Publisher = {Prentice Hall},
	Timestamp = {2015.04.16},
	Title = {Network flows: Theory, algorithms and applications},
	Year = {1993}}

@article{alahicvpr2014,
	Author = {A. Alahi and V. Ramanathan and and L. Fei-Fei},
	Journal = {CVPR},
	Owner = {laurale},
	Timestamp = {2015.02.05},
	Title = {Socially-aware large-scale crowd forecasting},
	Year = {2014}}

@article{alieccv2008,
	Author = {S. Ali and M. Shah},
	Date-Added = {2011-01-19 14:12:20 +0100},
	Date-Modified = {2011-01-19 14:12:51 +0100},
	Journal = {ECCV},
	Owner = {laurale},
	Timestamp = {2015.04.16},
	Title = {Floor fields for tracking in high density crowded scenes},
	Year = {2008}}

@article{amereccv2012,
	Author = {M.R. Amer and D. Xie and M. Zhao and S. Todorovic and S.-C. Zhu},
	Date-Added = {2012-11-01 19:50:30 +0000},
	Date-Modified = {2012-11-01 19:51:24 +0000},
	Journal = {ECCV},
	Owner = {laurale},
	Timestamp = {2015.04.16},
	Title = {Cost-sensitive top-down/bottom-up inference for multiscale activity recognition},
	Year = {2012}}

@article{andrilukacvpr2008,
	Author = {M. Andriluka and S. Roth and B. Schiele},
	Date-Added = {2010-12-01 17:07:12 +0100},
	Date-Modified = {2010-12-01 17:07:57 +0100},
	Journal = {CVPR},
	Owner = {laurale},
	Timestamp = {2015.04.16},
	Title = {People-tracking-by-detection and people-detection-by-tracking},
	Year = {2008}}

@article{andriyenkocvpr2011,
	Author = {A. Andriyenko and K. Schindler},
	Date-Added = {2014-03-06 13:26:37 +0000},
	Date-Modified = {2014-03-06 13:29:11 +0000},
	Journal = {CVPR},
	Rss-Description = {@inproceedings{Andriyenko:2011:MTT, Author = {Anton Andriyenko and Konrad Schindler}, Booktitle = {CVPR}, Title = {Multi-target Tracking by Continuous Energy Minimization}, Year = {2011} }
},
	Title = {Discrete-continuous optimization for multi-target tracking},
	Year = {2011}}

@article{andriyenkocvpr2012,
	Author = {A. Andriyenko and K. Schindler and S. Roth},
	Date-Added = {2013-04-04 15:35:12 +0000},
	Date-Modified = {2013-04-04 15:35:48 +0000},
	Journal = {CVPR},
	Title = {Discrete-Continuous Optimization for Multi-Target Tracking},
	Year = {2012}}

@article{arbelaezcvpr2014,
	Author = {P. Arbel\'aez and J. Pont-Tuset and J.T. Barron and F. Marqu\'es and J. Malik},
	Journal = {CVPR},
	Owner = {laurale},
	Timestamp = {2015.04.21},
	Title = {Multiscale Combinatorial Grouping},
	Year = {2014}}

@article{baecvpr2014,
	Author = {S. Bae and K. Yoon},
	Journal = {CVPR},
	Owner = {laurale},
	Timestamp = {2016.03.07},
	Title = {Robust Online Multi-Object Tracking based on Tracklet Confidence and Online Discriminative Appearance Learning},
	Year = {2014}}

@article{branchandprice,
	Author = {C. Barnhart and E.L. Johnson and G.L. Nemhauser and M.W.P. Savelsbergh and P.H. Vance},
	Date-Added = {2011-10-31 20:03:40 +0100},
	Date-Modified = {2011-10-31 20:08:53 +0100},
	Journal = {Operations Research},
	Owner = {laurale},
	Timestamp = {2015.04.16},
	Title = {Branch-and-price: column generation for solving huge integer programs},
	Volume = {46},
	Year = {1996}}

@article{benfoldcvpr2011,
	Author = {B. Benfold and I. Reid},
	Date-Added = {2011-07-06 16:58:33 +0200},
	Date-Modified = {2011-07-06 16:58:49 +0200},
	Journal = {CVPR},
	Owner = {laurale},
	Timestamp = {2015.04.16},
	Title = {Stable multi-target tracking in real-time surveillance video},
	Year = {2011}}

@article{berclazcvpr2006,
	Author = {J. Berclaz and F. Fleuret and P. Fua},
	Date-Added = {2010-12-01 16:04:36 +0100},
	Date-Modified = {2011-01-31 14:23:39 +0100},
	Journal = {CVPR},
	Owner = {laurale},
	Timestamp = {2015.04.16},
	Title = {Robust people tracking with global trajectory optimization},
	Year = {2006}}

@article{berclazpets2009,
	Author = {J. Berclaz and F. Fleuret and P. Fua},
	Date-Added = {2010-10-21 17:54:35 +0200},
	Date-Modified = {2011-02-27 14:58:49 +0100},
	Journal = {12th IEEE International Workshop on PETS},
	Owner = {laurale},
	Timestamp = {2015.04.16},
	Title = {Multiple object tracking using flow linear programming},
	Year = {2009}}

@article{berclaztpami2011,
	Author = {J. Berclaz and F. Fleuret and E. T{\"u}retken and P. Fua},
	Date-Added = {2011-02-27 14:58:51 +0100},
	Date-Modified = {2011-02-27 14:59:21 +0100},
	Journal = {TPAMI},
	Owner = {laurale},
	Timestamp = {2015.04.16},
	Title = {Multiple object tracking using k-shortest paths optimization},
	Year = {2011}}

@book{DD,
	Author = {D. Bertsekas},
	Date-Added = {2011-11-20 18:33:18 +0100},
	Date-Modified = {2011-11-20 18:34:03 +0100},
	Owner = {laurale},
	Publisher = {Athena Scientific},
	Timestamp = {2015.04.16},
	Title = {Nonlinear programming},
	Year = {1999}}

@article{bewleyicra2016,
	Author = {A. Bewley and L. Ott and F. Ramos and B. Upcroft},
	Journal = {ICRA},
	Owner = {laurale},
	Timestamp = {2016.03.07},
	Title = {ALExTRAC: Affinity Learning by Exploring Temporal Reinforcement within Association Chains},
	Year = {2016}}

@article{ctm,
	Author = {D.M. Blei and J.D. Lafferty},
	Date-Added = {2012-11-02 01:16:51 +0000},
	Date-Modified = {2012-11-02 01:18:04 +0000},
	Journal = {The Annals of Applied Statistics},
	Number = {1},
	Owner = {laurale},
	Timestamp = {2015.04.16},
	Title = {A correlated topic model of SCIENCE},
	Volume = {1},
	Year = {2007}}

@article{rf,
	Author = {L. Breiman},
	Date-Added = {2012-11-02 16:15:38 +0000},
	Date-Modified = {2012-11-02 16:16:04 +0000},
	Journal = {Machine Learning},
	Number = {1},
	Pages = {5-32},
	Title = {Random forests},
	Volume = {45},
	Year = {2001}}

@article{breitensteiniccv2009,
	Author = {M.D. Breitenstein and F. Reichlin and B. Leibe and E. Koller-Meier and L. van Gool},
	Date-Added = {2010-12-01 17:08:03 +0100},
	Date-Modified = {2011-01-31 14:24:51 +0100},
	Journal = {ICCV},
	Owner = {laurale},
	Timestamp = {2015.04.16},
	Title = {Robust tracking-by-detection using a detector confidence particle filter},
	Year = {2009}}

@article{brendelcvpr2011,
	Author = {W. Brendel and M.R. Amer and S. Todorovic},
	Journal = {CVPR},
	Owner = {lealtaix},
	Timestamp = {2016.07.19},
	Title = {Multiobject tracking as maximum weight indepen- dent set},
	Year = {2011}}

@article{brostowcvpr2006,
	Author = {G.J. Brostow and R. Cipolla},
	Date-Added = {2011-07-08 11:50:04 +0200},
	Date-Modified = {2011-07-08 11:50:46 +0200},
	Journal = {CVPR},
	Owner = {laurale},
	Timestamp = {2015.04.16},
	Title = {Unsupervised detection of independent motion in crowds},
	Year = {2006}}

@article{browstowcvpr2006,
	Author = {G. Browstow and R. Cipolla},
	Date-Added = {2011-01-19 14:12:54 +0100},
	Date-Modified = {2011-01-19 14:14:01 +0100},
	Journal = {CVPR},
	Owner = {laurale},
	Timestamp = {2015.04.16},
	Title = {Unsupervised bayesian detection of independent motion in crowds},
	Year = {2006}}

@article{broxeccv2010,
	Author = {T. Brox and J. Malik},
	Date-Added = {2014-11-09 17:08:43 +0000},
	Date-Modified = {2014-11-09 17:09:25 +0000},
	Journal = {ECCV},
	Owner = {laurale},
	Timestamp = {2015.04.16},
	Title = {Object Segmentation by Long Term Analysis of Point Trajectories},
	Year = {2010}}

@article{buttcvpr2013,
	Author = {A. Butt and R. Collins},
	Date-Added = {2014-03-06 13:06:26 +0000},
	Date-Modified = {2014-03-06 13:07:01 +0000},
	Journal = {CVPR},
	Owner = {laurale},
	Timestamp = {2015.04.16},
	Title = {Multi-target Tracking by {L}agrangian Relaxation to Min-Cost Network Flow},
	Year = {2013}}

@article{chardaire1995,
	Author = {P. Chardaire and A. Sutter},
	Date-Added = {2011-11-19 13:02:36 +0100},
	Date-Modified = {2011-11-19 13:02:36 +0100},
	Journal = {Management Science},
	Owner = {laurale},
	Timestamp = {2015.04.16},
	Title = {A decomposition method for quadratic zero-one programming},
	Year = {1995}}

@article{charicvpr2015,
	Author = {V. Chari and S. Lacoste-Julien and I. Laptev and J. Sivic},
	Date-Added = {2016-03-11 15:03:41 +0000},
	Date-Modified = {2016-03-11 15:04:18 +0000},
	Journal = {CVPR},
	Title = {On pairwise costs for network flow multi-object tracking},
	Year = {2015}}

@article{chencvpr2014,
	Author = {S. Chen and A. Fern and S. Todorovic},
	Journal = {CVPR},
	Owner = {lealtaix},
	Timestamp = {2016.07.19},
	Title = {Multi-object tracking via constrained sequential labeling},
	Year = {2014}}

@misc{chen2015xgboost,
	Author = {Chen, Tianqi and He, Tong},
	Howpublished = {\href{https://github.com/dmlc/xgboost}{GitHub}},
	Title = {xgboost: eXtreme Gradient Boosting},
	Year = {2015}}

@article{Chen2015,
	Author = {Y. Chen and X. Yang and B. Zhong and S. Pan and D. Chen and H. Zhang},
	Journal = {Applied Soft Computing},
	Title = {{CNNTracker}: Online discriminative object tracking via deep convolutional neural network},
	Year = {2015}}

@article{choiiccv2015,
	Author = {W. Choi},
	Date-Modified = {2015-11-04 13:58:29 +0000},
	Journal = {ICCV},
	Owner = {laurale},
	Title = {Near-Online Multi-target Tracking with Aggregated Local Flow Descriptor},
	Year = {2015}}

@article{choieccv2010,
	Author = {W. Choi and S. Savarese},
	Date-Added = {2011-09-03 19:53:17 +0200},
	Date-Modified = {2011-09-03 19:54:10 +0200},
	Journal = {ECCV},
	Owner = {laurale},
	Timestamp = {2015.04.16},
	Title = {Multiple target tracking in world coordinate with single, minimally calibrated camera},
	Year = {2010}}

@article{choieccv2012,
	Author = {W. Choi and S. Savarese},
	Date-Added = {2012-11-01 19:35:24 +0000},
	Date-Modified = {2012-11-01 19:45:15 +0000},
	Journal = {ECCV},
	Owner = {laurale},
	Timestamp = {2015.04.16},
	Title = {A unified framework for multi-target tracking and collective activity recognition},
	Year = {2012}}

@article{choicvpr2011,
	Author = {W. Choi and K. Shahid and S. Savarese},
	Date-Added = {2012-11-01 19:47:42 +0000},
	Date-Modified = {2012-11-01 19:48:14 +0000},
	Journal = {CVPR},
	Owner = {laurale},
	Timestamp = {2015.04.16},
	Title = {Learning context for collective activity recognition},
	Year = {2011}}

@article{choiiccvw2009,
	Author = {W. Choi and K. Shahid and S. Savarese},
	Date-Added = {2012-11-01 19:46:27 +0000},
	Date-Modified = {2012-11-01 19:47:39 +0000},
	Journal = {ICCV. 9th International Workshop on Visual Surveillance (VSWS)},
	Owner = {laurale},
	Timestamp = {2015.04.16},
	Title = {What are they doing?: Collective activity classification using spatio-temporal relationship among people},
	Year = {2009}}

@article{chopra2006,
	Author = {Chopra, S. and Hadsell, R. and LeCun, Y.},
	Journal = {CVPR},
	Title = {Learning a similarity metric discriminatively, with application to face verification},
	Year = {2005}}

@article{RFtechnicalreport,
	Author = {A. Criminisi and J. Shotton and E. Konukoglu},
	Date-Added = {2012-11-07 22:36:18 +0000},
	Date-Modified = {2012-11-07 22:39:39 +0000},
	Journal = {MSR-TR-2011-114},
	Title = {Decision forests for classification, regression, density estimation, manifold learning and semi-supervised learning},
	Year = {2011}}

@article{dalalcvpr2005,
	Author = {N. Dalal and B. Triggs},
	Date-Added = {2010-12-01 17:09:24 +0100},
	Date-Modified = {2010-12-01 17:10:11 +0100},
	Journal = {CVPR},
	Owner = {laurale},
	Timestamp = {2015.04.16},
	Title = {Histograms of oriented gradients for human detection},
	Year = {2005}}

@article{hog,
	Author = {N. Dalal and B. Triggs and C. Schmid},
	Date-Added = {2012-11-02 15:50:19 +0000},
	Date-Modified = {2012-11-14 16:47:37 +0000},
	Journal = {ECCV},
	Title = {Human detection using oriented histograms of flow and appearance},
	Year = {2006}}

@book{LP,
	Author = {G.B. Dantzig},
	Date-Added = {2011-02-17 18:22:54 +0100},
	Date-Modified = {2011-02-17 18:23:44 +0100},
	Owner = {laurale},
	Publisher = {Princeton University Press, Princenton, NJ},
	Timestamp = {2015.04.16},
	Title = {Linear programming and extensions},
	Year = {1963}}

@article{deb1997,
	Author = {S. Deb and M. Yeddanapuddi and K. Pattipati and Y.Bar-Shalom},
	Date-Added = {2011-11-21 16:50:42 +0100},
	Date-Modified = {2011-11-21 16:52:18 +0100},
	Journal = {IEEE Trans. on AES},
	Owner = {laurale},
	Timestamp = {2015.04.16},
	Title = {A generalized s-d assignment algorithm for multisensor-multitarget state estimation},
	Year = {1997}}

@article{dehghancvpr2015,
	Author = {A. Dehghan and S.M. Assari and M. Shah},
	Journal = {CVPR},
	Owner = {lealtaix},
	Timestamp = {2016.07.19},
	Title = {GMMCP-tracker: Globally optimal generalized maximum multi clique problem for multiple object tracking},
	Year = {2015}}

@article{dicleiccv2013,
	Author = {C. Dicle and O. Camps and M. Sznaier},
	Journal = {ICCV},
	Owner = {laurale},
	Timestamp = {2016.03.07},
	Title = {The Way They Move: Tracking Targets with Similar Appearance},
	Year = {2013}}

@misc{DielemanPlankton2015,
	Author = {Sander Dieleman and A\"aron van den Oord and Ira Korshunova and Jeroen Burms and Jonas Degrave and Lionel Pigou and Pieter Buteneers},
	Howpublished = {\href{http://benanne.github.io/2015/03/17/plankton.html}{Blog entry}},
	Title = {Classifying plankton with deep neural networks},
	Year = {2015}}

@article{Dieleman2015,
	Author = {S. Dieleman and K.W. Willett and J. Dambre},
	Journal = {Monthly Notices of the Royal Astronomical Society},
	Number = {10},
	Pages = {1441--1459},
	Title = {Rotation-invariant convolutional neural networks for galaxy morphology prediction},
	Volume = {450},
	Year = {2015}}

@article{caltechpedestrians,
	Author = {P. Doll{\'a}r and C. Wojek and B. Schiele and P. Perona},
	Date-Added = {2014-04-07 16:15:27 +0000},
	Date-Modified = {2014-04-07 16:16:29 +0000},
	Journal = {CVPR},
	Owner = {laurale},
	Timestamp = {2015.02.03},
	Title = {Pedestrian Detection: A Benchmark},
	Year = {2009}}

@article{dollarpami2014,
	Author = {Piotr Doll\'ar and Ron Appel and Serge Belongie and Pietro Perona},
	Journal = {PAMI},
	Title = {Fast Feature Pyramids for Object Detection},
	Year = {2014}}

@article{dollartpami2014,
	Author = {P. Dol{\'l}ar and R. Appel and S. Belongie and P. Perona},
	Journal = {TPAMI},
	Owner = {lealtaix},
	Timestamp = {2016.07.19},
	Title = {Fast feature pyramids for object detection},
	Year = {2014}}

@article{essiccv2007,
	Author = {A. Ess and B. Leibe and L. van Gool},
	Date-Added = {2013-10-25 18:04:24 +0000},
	Date-Modified = {2013-10-25 18:04:51 +0000},
	Journal = {ICCV},
	Title = {Depth and appearance for mobile scene analysis},
	Year = {2007}}

@article{esscvpr2008,
	Author = {A. Ess and B. Leibe and K. Schindler and L. van Gool},
	Date-Added = {2013-10-25 18:05:50 +0000},
	Date-Modified = {2013-10-25 18:06:25 +0000},
	Journal = {CVPR},
	Title = {A mobile vision system for robust multi-person tracking},
	Year = {2008}}

@article{fan2010human,
	Author = {J. Fan and W. Xu and Y. Wu and Y. Gong},
	Journal = {IEEE Transactions on Neural Networks},
	Number = {10},
	Pages = {1610--1623},
	Title = {Human tracking using convolutional neural networks},
	Volume = {21},
	Year = {2010}}

@article{partbased,
	Author = {P.F. Felzenszwalb and R.B. Girshick and D. McAllester and D. Ramanan},
	Date-Added = {2012-11-01 20:10:34 +0000},
	Date-Modified = {2012-11-01 20:11:32 +0000},
	Journal = {TPAMI},
	Owner = {laurale},
	Timestamp = {2015.04.16},
	Title = {Object Detection with Discriminatively Trained Part Based Models},
	Year = {2010}}

@article{felzenszwalbijcv2006,
	Author = {Pedro F. Felzenszwalb and Daniel P. Huttenlocher},
	Journal = {IJCV},
	Owner = {laurale},
	Timestamp = {2015.03.02},
	Title = {Efficient Belief Propagation for Early Vision},
	Year = {2006}}

@article{pets2009,
	Author = {J.M. Ferryman},
	Date-Added = {2011-02-17 19:43:15 +0100},
	Date-Modified = {2011-02-17 19:44:06 +0100},
	Owner = {laurale},
	Timestamp = {2015.04.16},
	Title = {PETS 2009 dataset: Performance and evaluation of tracking and surveillance},
	Year = {2009}}

@article{flownet2015,
	Author = {P. Fischer and A. Dosovitskiy and E. Ilg and P. H\"ausser and C. Hazirbas and V. Golkov and P. van der Smagt and D. Cremers and T. Brox},
	Journal = {ICCV},
	Title = {{FlowNet: Learning Optical Flow with Convolutional Networks}},
	Year = {2015}}

@article{fleurettpami2008,
	Author = {F. Fleuret and J. Berclaz and R. Lengagne and P. Fua},
	Date-Added = {2011-11-19 13:02:29 +0100},
	Date-Modified = {2011-11-19 13:02:29 +0100},
	Journal = {TPAMI},
	Owner = {laurale},
	Timestamp = {2015.04.16},
	Title = {Multi-camera people tracking with a probabilistic occupancy map},
	Year = {2008}}

@article{deepStereo2015,
	Author = {J. Flynn and I. Neulander and J. Philbin and N. Snavely},
	Journal = {arXiv:1506.06825},
	Owner = {laurale},
	Timestamp = {2015.04.17},
	Title = {{DeepStereo}: Learning to Predict New Views from the World's Imagery},
	Year = {2015}}

@article{fragkiadakicvpr2011,
	Author = {K. Fragkiadaki and J. Shi},
	Journal = {CVPR},
	Owner = {lealtaix},
	Timestamp = {2016.07.19},
	Title = {Detection free tracking: Exploiting motion and topology for segmenting and tracking under entanglement},
	Year = {2011}}

@article{fragkiadakieccv2012,
	Author = {K. Fragkiadaki and W. Zhang and G. Zhng and J. Shi},
	Date-Added = {2014-11-09 17:25:02 +0000},
	Date-Modified = {2014-11-09 17:26:04 +0000},
	Journal = {ECCV},
	Title = {Two-granularity tracking: mediating trajectory and detections graphs for tracking under occlusions},
	Year = {2012}}

@article{Friedman2002GBM,
	Author = {J.H. Friedman},
	Journal = {Computational Statistics \& Data Analysis},
	Number = {4},
	Pages = {367--378},
	Title = {Stochastic gradient boosting},
	Volume = {38},
	Year = {2002}}

@article{galltpami2011,
	Author = {J. Gall and A. Yao and N. Razavi and L. van Gool and V. Lempitsky},
	Date-Added = {2012-11-01 19:51:42 +0000},
	Date-Modified = {2013-04-04 08:51:34 +0000},
	Journal = {TPAMI},
	Owner = {laurale},
	Timestamp = {2015.04.16},
	Title = {Hough forests for object detection, tracking and action recognition},
	Year = {2011}}

@article{gewacv2009,
	Author = {W. Ge and R.T. Collins and B. Ruback},
	Date-Added = {2011-09-03 19:48:40 +0200},
	Date-Modified = {2011-09-03 19:49:15 +0200},
	Journal = {WACV},
	Owner = {laurale},
	Timestamp = {2015.04.16},
	Title = {Automatically detecting the small group structure of a crowd},
	Year = {2009}}

@article{geigerpami2014,
	Author = {A. Geiger and M. Lauer and C. Wojek and C. Stiller and R. Urtasun},
	Journal = {TPAMI},
	Owner = {laurale},
	Timestamp = {2016.03.07},
	Title = {3D Traffic Scene Understanding from Movable Platforms},
	Year = {2014}}

@article{shucvpr2013,
	Author = {Guang Shu, Afshin Dehghan and Mubarak Shah},
	Date-Added = {2015-11-03 16:22:12 +0000},
	Date-Modified = {2015-11-03 16:23:43 +0000},
	Journal = {CVPR},
	Title = {Improving an Object Detector and Extracting Regions using Superpixels},
	Year = {2013}}

@article{kaiming2015,
	Author = {Kaiming He and Xiangyu Zhang and Shaoqing Ren and Jian Sun},
	Journal = {ICCV},
	Title = {Delving Deep into Rectifiers: Surpassing Human-Level Performance on ImageNet Classification},
	Year = {2015}}

@article{SFM,
	Author = {D. Helbing and P. Moln{\'a}r},
	Date-Added = {2011-02-17 16:05:52 +0100},
	Date-Modified = {2011-02-17 16:07:27 +0100},
	Journal = {Physical Review E},
	Owner = {laurale},
	Pages = {4282},
	Timestamp = {2015.04.16},
	Title = {Social force model for pedestrian dynamics},
	Volume = {51},
	Year = {1995}}

@techreport{huangtech2007,
	Author = {G. Huang and M. Ramesh and T. BerT. Berg. Learned-Miller},
	Institution = {University of Massachussetts, Amherst},
	Number = {07-49},
	Owner = {laurale},
	Timestamp = {2015.02.13},
	Title = {Labeled face in the wild: a datadata for studying face recognition in unconstrained environments.},
	Year = {2007}}

@article{izadiniaeccv2012,
	Author = {H. Izadinia and I. Saleemi and W. Li and M. Shah},
	Journal = {ECCV},
	Owner = {lealtaix},
	Timestamp = {2016.07.19},
	Title = {(MP)2T: Multiple people multiple parts tracker},
	Year = {2011}}

@article{jia2014caffe,
	Author = {Jia, Yangqing and Shelhamer, Evan and Donahue, Jeff and Karayev, Sergey and Long, Jonathan and Girshick, Ross and Guadarrama, Sergio and Darrell, Trevor},
	Journal = {arXiv preprint arXiv:1408.5093},
	Title = {Caffe: Convolutional Architecture for Fast Feature Embedding},
	Year = {2014}}

@article{jiangcvpr2007,
	Author = {H. Jiang and S. Fels and J.J. Little},
	Date-Added = {2010-10-21 17:53:13 +0200},
	Date-Modified = {2011-01-19 14:15:12 +0100},
	Journal = {CVPR},
	Owner = {laurale},
	Timestamp = {2015.04.16},
	Title = {A linear programming approach for multiple object tracking},
	Year = {2007}}

@article{SFM2,
	Author = {A. Johansson and D. Helbing and P. Shukla},
	Date-Added = {2011-02-17 16:07:35 +0100},
	Date-Modified = {2012-11-02 16:17:32 +0000},
	Journal = {Advances in complex systems},
	Owner = {laurale},
	Timestamp = {2015.04.16},
	Title = {Specification of a microscopic pedestrian model by evolutionary adjustment to video tracking data},
	Year = {2007}}

@article{clear,
	Author = {R. Kasturi and D. Goldgof and P. Soundararajan and V. Manohar and J. Garofolo and M. Boonstra and V. Korzhova and J. Zhang},
	Date-Added = {2011-02-10 19:04:59 +0100},
	Date-Modified = {2011-03-01 16:29:50 +0100},
	Journal = {TPAMI},
	Number = {2},
	Owner = {laurale},
	Timestamp = {2015.04.16},
	Title = {Framework for performance evaluation for face, text and vehicle detection and tracking in video: data, metrics, and protocol},
	Volume = {31},
	Year = {2009}}

@article{kauciccvpr2005,
	Author = {R. Kaucic and A.G. Perera and G. Brooksby and J. Kaufhold and A. Hoogs},
	Date-Added = {2010-12-01 17:02:39 +0100},
	Date-Modified = {2010-12-01 17:03:34 +0100},
	Journal = {CVPR},
	Owner = {laurale},
	Timestamp = {2015.04.16},
	Title = {A unified framework for tracking through occlusions and across sensor gaps},
	Year = {2005}}

@article{khamiscvpr2012,
	Author = {S. Khamis and V.I. Morariu and L.S. Davis},
	Date-Added = {2012-11-01 19:49:48 +0000},
	Date-Modified = {2012-11-01 19:50:21 +0000},
	Journal = {CVPR},
	Owner = {laurale},
	Timestamp = {2015.04.16},
	Title = {A flow model for joint action recognition and identity maintenance},
	Year = {2012}}

@article{khantpami2005,
	Author = {Z. Khan and T. Balch and F. Dellaert},
	Date-Added = {2010-12-01 16:28:36 +0100},
	Date-Modified = {2011-03-01 16:30:15 +0100},
	Journal = {TPAMI},
	Owner = {laurale},
	Timestamp = {2015.04.16},
	Title = {MCMC-based particle filtering for tracking a variable number of interacting targets},
	Year = {2005}}

@article{kimiccv2015,
	Author = {Chanho Kim and Fuxin Li and Arridhana Ciptadi and James Rehg},
	Date-Added = {2015-11-03 16:36:32 +0000},
	Date-Modified = {2015-11-03 16:37:10 +0000},
	Journal = {ICCV},
	Title = {Multiple Hypothesis Tracking Revisited: Blending in Modern Appearance Model},
	Year = {2015}}

@article{kitanieccv2012,
	Author = {K.M Kitani and B.D. Ziebart and J.A. Bagnell and M. Hebert},
	Date-Added = {2012-11-01 19:45:20 +0000},
	Date-Modified = {2012-11-01 19:45:55 +0000},
	Journal = {ECCV},
	Owner = {laurale},
	Timestamp = {2015.04.16},
	Title = {Activity forecasting},
	Year = {2012}}

@article{komodakisiccv2007,
	Author = {N. Komodakis and N. Paragios and G. Tziritas},
	Date-Added = {2011-11-21 16:18:09 +0100},
	Date-Modified = {2011-11-21 16:18:52 +0100},
	Journal = {ICCV},
	Owner = {laurale},
	Timestamp = {2015.04.16},
	Title = {MRF optimization via Dual Decomposition: Message-Passing Revisited},
	Year = {2007}}

@article{VOC2014,
	Author = {Matej Kristan and others},
	Journal = {European Conference on Computer Vision Workshops (ECCVW). Visual Object Tracking Challenge Workshop},
	Owner = {laurale},
	Timestamp = {2015.02.03},
	Title = {The Visual Object Tracking VOT2014 challenge results},
	Year = {2014}}

@article{krizhevskyImageNet,
	Author = {A. Krizhevsky and I. Sutskever and G. Hinton},
	Journal = {ANIPS},
	Owner = {laurale},
	Timestamp = {2015.02.05},
	Title = {{ImageNet} classification with deep convolutional neural networks},
	Year = {2012}}

@article{kuocvpr2011,
	Author = {Cheng-Hao Kuo and Ram Nevatia},
	Date-Added = {2015-11-03 16:34:16 +0000},
	Date-Modified = {2015-11-03 16:34:35 +0000},
	Journal = {CVPR},
	Title = {How does Person Identity Recognition Help Multi-Person Tracking?},
	Year = {2011}}

@article{lancvpr2012,
	Author = {T. Lan and L. Sigal and G. Mori},
	Date-Added = {2012-11-01 19:48:47 +0000},
	Date-Modified = {2012-11-01 19:49:17 +0000},
	Journal = {CVPR},
	Owner = {laurale},
	Timestamp = {2015.04.16},
	Title = {Social roles in hierarchical models for human activity recognition},
	Year = {2012}}

@article{lealcvpr2014,
	Author = {L. Leal-Taix{\'e} and M. Fenzi and A. Kuznetsova and B. Rosenhahn and S. Savarese},
	Date-Added = {2014-11-09 11:47:46 +0000},
	Date-Modified = {2014-11-09 11:48:45 +0000},
	Journal = {CVPR},
	Title = {Lerning an image-based motion context for multiple people tracking},
	Year = {2014}}

@article{MOTChallenge:arxiv:2015,
	Author = {Leal-Taix{\'e}, L. and Milan, A. and Reid, I. and Roth, S. and Schindler, K.},
	Journal = {arXiv:1504.01942},
	Owner = {laurale},
	Timestamp = {2015.04.12},
	Title = {MOTChallenge 2015: Towards a Benchmark for Multi-Target Tracking},
	Year = {2015}}

@article{lealcvpr2012,
	Author = {L. Leal-Taix{\'e} and G. Pons-Moll and B. Rosenhahn},
	Date-Added = {2012-10-28 20:59:23 +0000},
	Date-Modified = {2012-10-28 20:59:55 +0000},
	Journal = {CVPR},
	Owner = {laurale},
	Timestamp = {2015.04.16},
	Title = {Branch-and-price global optimization for multi-view multi-object tracking},
	Year = {2012}}

@article{lealiccv2011,
	Author = {L. Leal-Taix{\'e} and G. Pons-Moll and B. Rosenhahn},
	Date-Added = {2012-10-28 20:59:17 +0000},
	Date-Modified = {2012-10-28 20:59:17 +0000},
	Journal = {ICCV. 1st Workshop on Modeling, Simulation and Visual Analysis of Large Crowds},
	Owner = {laurale},
	Timestamp = {2015.04.16},
	Title = {Everybody needs somebody: Modeling social and grouping behavior on a linear programming multiple people tracker},
	Year = {2011}}

@article{leibetpami2008,
	Author = {B. Leibe and K. Schindler and N. Cornelis and L. van Gool},
	Date-Added = {2010-12-01 17:01:47 +0100},
	Date-Modified = {2010-12-01 17:02:33 +0100},
	Journal = {TPAMI},
	Number = {10},
	Owner = {laurale},
	Timestamp = {2015.04.16},
	Title = {Coupled detection and tracking from static cameras and moving vehicles},
	Volume = {30},
	Year = {2008}}

@article{leibeiccv2007,
	Author = {B. Leibe and K. Schindler and L. van Gool},
	Date-Added = {2011-02-11 15:59:14 +0100},
	Date-Modified = {2011-02-11 16:01:38 +0100},
	Journal = {ICCV},
	Owner = {laurale},
	Timestamp = {2015.04.16},
	Title = {Coupled detection and trajectory estimation from multi-object tracking},
	Year = {2007}}

@article{lernereurographics2007,
	Author = {A. Lerner and Y. Chrysanthou and D. Lischinski},
	Date-Added = {2011-02-17 16:10:33 +0100},
	Date-Modified = {2011-02-17 16:11:19 +0100},
	Journal = {Eurographics},
	Number = {3},
	Owner = {laurale},
	Timestamp = {2015.04.16},
	Title = {Crowds by Example},
	Volume = {26},
	Year = {2007}}

@article{Li2015,
	Archiveprefix = {arXiv},
	Author = {{Li}, H. and {Li}, Y. and {Porikli}, F.},
	Eprint = {1503.00072},
	Journal = {ArXiv e-prints},
	Title = {DeepTrack: Learning Discriminative Feature Representations Online for Robust Visual Tracking},
	Year = {2015}}

@article{limia2008,
	Author = {K. Li and E. Miller and M. Chen and T. Kanade and L.E. Weiss and P.G. Campbell},
	Date-Added = {2013-05-02 15:57:21 +0000},
	Date-Modified = {2013-05-02 15:57:21 +0000},
	Journal = {Medical Image Analysis},
	Month = {October},
	Number = {5},
	Owner = {laurale},
	Pages = {546-566},
	Timestamp = {2015.02.05},
	Title = {Cell population tracking and lineage construction with spatiotemporal context},
	Volume = {12},
	Year = {2008}}

@article{licvpr2009,
	Author = {Y. Li and C. Huang and R. Nevatia},
	Date-Added = {2013-10-22 15:00:31 +0000},
	Date-Modified = {2013-10-22 15:01:11 +0000},
	Journal = {CVPR},
	Title = {Learning to associate: hybrid boosted multi-target tracker for crowded scene},
	Year = {2009}}

@article{lubericra2010,
	Author = {M. Luber and J.A. Stork and G.D. Tipaldi and K.O Arras},
	Date-Added = {2011-02-17 16:33:57 +0100},
	Date-Modified = {2011-02-17 16:36:43 +0100},
	Journal = {ICRA},
	Owner = {laurale},
	Timestamp = {2015.04.16},
	Title = {People tracking with human motion predictions from social forces},
	Year = {2010}}

@article{glpk,
	Author = {A. Makhorin},
	Date-Added = {2011-02-17 18:25:55 +0100},
	Date-Modified = {2011-02-17 18:31:06 +0100},
	Journal = {http://www.gnu.org/software/glpk/},
	Owner = {laurale},
	Timestamp = {2015.04.16},
	Title = {GNU Linear Programming Kit (GLPK)},
	Year = {2010}}

@article{mclaughlinwacv2015,
	Author = {N. McLaughlin and J. Martinez Del Rincon and P. Miller},
	Journal = {WACV},
	Owner = {laurale},
	Timestamp = {2015.11.06},
	Title = {Enhancing Linear Programming with Motion Modeling for Multi-target Tracking.},
	Year = {2015}}

@article{mehrancvpr2009,
	Author = {R. Mehran and A. Oyama and M. Shah},
	Date-Added = {2010-10-21 17:58:47 +0200},
	Date-Modified = {2010-10-21 18:01:06 +0200},
	Journal = {CVPR},
	Owner = {laurale},
	Timestamp = {2015.04.16},
	Title = {Abnormal crowd behavior detection using social force model},
	Year = {2009}}

@conference{milancvpr2015,
	Author = {A. Milan and L. Leal-Taix{\'e} and K. Schindler and I. Reid},
	Booktitle = {CVPR},
	Owner = {laurale},
	Timestamp = {2015.04.17},
	Title = {Joint Tracking and Segmentation of Multiple Targets},
	Year = {2015}}

@article{milancvpr2013,
	Author = {A. Milan and K. Schindler and S. Roth},
	Date-Added = {2013-10-22 14:58:05 +0000},
	Date-Modified = {2013-10-22 14:58:46 +0000},
	Journal = {CVPR},
	Title = {Detection- and trajectory-level exclusion in multiple object tracking},
	Year = {2013}}

@article{milantpami2016,
	Author = {A. Milan and K. Schindler and S. Roth},
	Journal = {TPAMI},
	Owner = {laurale},
	Timestamp = {2016.03.07},
	Title = {Multi-Target Tracking by Discrete-Continuous Energy Minimization},
	Year = {2016}}

@article{milantpami2014,
	Author = {A Milan and S. RotS. Roth. Schindler},
	Journal = {TPAMI},
	Owner = {laurale},
	Timestamp = {2016.03.07},
	Title = {Continuous Energy Minimization for Multitarget Tracking},
	Year = {2014}}

@article{nips2001,
	Author = {A.Y. Ng and M. Jordan},
	Date-Added = {2012-11-03 02:31:36 +0000},
	Date-Modified = {2012-11-03 02:32:00 +0000},
	Journal = {NIPS},
	Title = {On Discriminative vs. Generative Classifiers: A comparison of logistic regression and Naive Bayes},
	Year = {2001}}

@article{nieblesijcv2008,
	Author = {J.C. Niebles and H. Wand and L. Fei-Fei},
	Date-Added = {2012-11-01 20:29:07 +0000},
	Date-Modified = {2012-11-01 20:29:51 +0000},
	Journal = {IJCV},
	Owner = {laurale},
	Timestamp = {2015.04.16},
	Title = {Unsupervised learning of human action categories using spatial-temporal words},
	Year = {2008}}

@article{nilliuscvpr2006,
	Author = {P. Nillius and J. Sullivan and S. Carlsson},
	Date-Added = {2010-12-01 16:17:40 +0100},
	Date-Modified = {2011-01-31 15:13:28 +0100},
	Journal = {CVPR},
	Owner = {laurale},
	Timestamp = {2015.04.16},
	Title = {Multi-target tracking - linking identities using bayesian network inference},
	Year = {2006}}

@article{okumaeccv2004,
	Author = {Kenji Okuma and Ali Taleghani and Nando De Freitas and James J. Little and David G. Lowe},
	Journal = {ECCV},
	Owner = {laurale},
	Timestamp = {2015.02.05},
	Title = {A Boosted Particle Filter: Multitarget Detection and Tracking},
	Year = {2004}}

@article{pelechanoeurographics2007,
	Author = {N. Pelechano and J.M. Allbeck and N.I. Badler},
	Date-Added = {2011-02-17 16:20:33 +0100},
	Date-Modified = {2011-02-17 16:21:35 +0100},
	Journal = {Eurographics/ACM SIGGRAPH Symposium on Computer Animation},
	Owner = {laurale},
	Timestamp = {2015.04.16},
	Title = {Controlling individual agents in high-density crowd simulation},
	Year = {2007}}

@article{pellegrinieccv2010,
	Author = {S. Pellegrini and A. Ess and L. van Gool},
	Date-Added = {2010-10-21 17:24:35 +0200},
	Date-Modified = {2010-10-21 18:01:12 +0200},
	Journal = {ECCV},
	Owner = {laurale},
	Timestamp = {2015.04.16},
	Title = {Improving data association by joint modeling of pedestrian trajectories and groupings},
	Year = {2010}}

@article{pellegriniiccv2009,
	Author = {S. Pellegrini and A. Ess and K. Schindler and L. van Gool},
	Date-Added = {2010-10-21 17:57:12 +0200},
	Date-Modified = {2010-10-21 18:01:19 +0200},
	Journal = {ICCV},
	Owner = {laurale},
	Timestamp = {2015.04.16},
	Title = {You'll never walk alone: modeling social behavior for multi-target tracking},
	Year = {2009}}

@article{pirsiavashcvpr2011,
	Author = {H. Pirsiavash and D. Ramanan and C.C. Fowlkes},
	Date-Added = {2011-09-03 19:41:03 +0200},
	Date-Modified = {2011-09-03 19:41:03 +0200},
	Journal = {CVPR},
	Owner = {laurale},
	Timestamp = {2015.04.16},
	Title = {Globally-optimal greedy algorithms for tracking a variable number of objects},
	Year = {2011}}

@article{poore1994,
	Author = {A.B. Poore},
	Date-Added = {2011-11-21 16:45:20 +0100},
	Date-Modified = {2011-11-21 16:46:05 +0100},
	Journal = {Computational Optimization and Applications},
	Owner = {laurale},
	Timestamp = {2015.04.16},
	Title = {Multidimensional assignment formulation of data association problems rising from multitarget and multisensor tracking},
	Year = {1994}}

@article{rezatofighiiccv2015,
	Author = {H. Rezatofighi and A. Milan and Z. Zhang and Q. Shi and A. Dick, I. Reid},
	Journal = {ICCV},
	Owner = {laurale},
	Timestamp = {2015.11.06},
	Title = {Joint Probabilistic Data Association Revisited},
	Year = {2015}}

@other{Riosa,
	Author = {J. Rios},
	Date-Added = {2011-11-20 13:37:09 +0100},
	Date-Modified = {2011-11-20 13:42:08 +0100},
	Title = {http://sourceforge.net/apps/wordpress/dwsolver/},
	Url = {http://sourceforge.net/apps/wordpress/dwsolver/},
	Urldate = {http://sourceforge.net/apps/wordpress/dwsolver/},
	Bdsk-Url-1 = {http://sourceforge.net/apps/wordpress/dwsolver/}}

@article{dantzigwolfe,
	Author = {J. Rios and K. Ross},
	Date-Added = {2011-10-31 20:09:09 +0100},
	Date-Modified = {2011-11-01 19:58:17 +0100},
	Journal = {Journal of Aerospace Computing, Information, and Communication},
	Number = {1},
	Owner = {laurale},
	Timestamp = {2015.04.16},
	Title = {Massively Parallel Dantzig-Wolfe Decomposition Applied to Traffic Flow Scheduling},
	Volume = {7},
	Year = {2010}}

@article{rodrigueziccv2009,
	Author = {M. Rodriguez and S. Ali and T. Kanade},
	Date-Added = {2012-11-02 01:18:09 +0000},
	Date-Modified = {2012-11-02 01:18:41 +0000},
	Journal = {ICCV},
	Title = {Tracking in unstructured crowded scenes},
	Year = {2009}}

@article{sadeghieccv2014,
	Author = {M.A. Sadeghi and D. Forsyth},
	Journal = {ECCV},
	Owner = {lealtaix},
	Timestamp = {2016.07.19},
	Title = {30Hz object detection with DPM V5},
	Year = {2014}}

@article{scovannericcv2009,
	Author = {P. Scovanner and M.F. Tappen},
	Date-Added = {2011-02-17 16:12:24 +0100},
	Date-Modified = {2011-02-17 16:12:51 +0100},
	Journal = {ICCV},
	Owner = {laurale},
	Timestamp = {2015.04.16},
	Title = {Learning pedestrian dynamics from the real world},
	Year = {2009}}

@article{shitriticcv2011,
	Author = {H.B. Shitrit and J. Berclaz and F. Fleuret and P. Fua},
	Date-Added = {2011-11-09 08:43:37 +0100},
	Date-Modified = {2011-11-09 08:45:59 +0100},
	Journal = {ICCV},
	Owner = {laurale},
	Timestamp = {2015.04.16},
	Title = {Tracking multiple people under global appearance constraints},
	Year = {2011}}

@article{shottoncvpr2011,
	Author = {J. Shotton and A. Fitzgibbon and M. Cook and T. Sharp and M. Finocchio and R. Moore and A. Kipman and A. Blake},
	Date-Added = {2012-11-01 20:03:08 +0000},
	Date-Modified = {2012-11-01 20:04:41 +0000},
	Journal = {CVPR},
	Owner = {laurale},
	Timestamp = {2015.04.16},
	Title = {Real-time human pose recognition in parts from a single depth image},
	Year = {2011}}

@article{shucvpr2012,
	Author = {G. Shu and A. Dehghan and O. Oreifej and E. Hand and M. Shah},
	Date-Added = {2013-04-11 19:05:22 +0000},
	Date-Modified = {2013-04-11 19:06:26 +0000},
	Journal = {CVPR},
	Title = {Part-based multiple-person tracking with partial occlusion handling},
	Year = {2012}}

@article{sigal2010humaneva,
	Author = {Sigal, L. and Balan, A.O. and Black, M.J.},
	Issn = {0920-5691},
	Journal = {IJCV},
	Number = {1},
	Owner = {laurale},
	Pages = {4--27},
	Publisher = {Springer},
	Timestamp = {2015.04.16},
	Title = {{Humaneva: Synchronized video and motion capture dataset and baseline algorithm for evaluation of articulated human motion}},
	Volume = {87},
	Year = {2010}}

@article{soleraiccv2015,
	Author = {F. Solera and S. Calderara and R. Cucchiara},
	Journal = {ICCV},
	Owner = {laurale},
	Timestamp = {2016.03.07},
	Title = {Learning to Divide and Conquer for Online Multi-Target Tracking},
	Year = {2015}}

@article{deepFace2014,
	Author = {Y. Taigman and Y. Ming and M. Ranzato, and L. Wolf},
	Journal = {CVPR},
	Title = {{DeepFace}: Closing the Gap to Human-Level Performance in Face Verification},
	Year = {2014}}

@article{tangiccv2013,
	Author = {S. Tang and M. Andriluka and A. Milan and K. Schindler and S. Roth and B. Schiele},
	Journal = {ICCV},
	Owner = {lealtaix},
	Timestamp = {2016.07.19},
	Title = {Learning people detectors for tracking in crowded scenes},
	Year = {2013}}

@techreport{tomasiklt1991,
	Author = {C. Tomasi and T. Kanade},
	Institution = {CMU-CS-91-132},
	Journal = {CMU-CS-91-132},
	Owner = {lealtaix},
	Timestamp = {2016.07.19},
	Title = {Detection and tracking of point features},
	Year = {1991}}

@article{torresanieccv2008,
	Author = {L. Torresani and V. Kolmogorov and C. Rother},
	Date-Added = {2011-11-19 13:02:34 +0100},
	Date-Modified = {2011-11-19 13:02:34 +0100},
	Journal = {ECCV},
	Owner = {laurale},
	Timestamp = {2015.04.16},
	Title = {Feature correspondence via graph matching: models and global optimization},
	Year = {2008}}

@article{walkcvpr2010,
	Author = {S. Walk and N. Majer and K. Schindler and B. Schiele},
	Date-Added = {2014-03-06 13:01:32 +0000},
	Date-Modified = {2014-03-06 13:02:03 +0000},
	Journal = {CVPR},
	Owner = {laurale},
	Timestamp = {2015.04.16},
	Title = {New Features and Insights for Pedestrian Detection},
	Year = {2010}}

@other{Wang2014,
	Author = {Wang, Haibo and Cruz-Roa, Angel and Basavanhally, Ajay and Gilmore, Hannah and Shih, Natalie and Feldman, Mike and Tomaszewski, John and Gonzalez, Fabio and Madabhushi, Anant},
	Journal = {Proc. SPIE},
	Title = {Cascaded ensemble of convolutional neural networks and handcrafted features for mitosis detection},
	Volume = {9041},
	Year = {2014}}

@article{Wang2015,
	Archiveprefix = {arXiv},
	Author = {N. Wang and S. Li and A. Gupta and D.-Y. Yeung},
	Eprint = {1501.04587},
	Journal = {ArXiv e-prints},
	Title = {Transferring Rich Feature Hierarchies for Robust Visual Tracking},
	Year = {2015}}

@article{wangbmvc2015,
	Author = {S. Wang and C. Fowlkes},
	Journal = {BMVC},
	Owner = {laurale},
	Timestamp = {2015.11.06},
	Title = {Learning Optimal Parameters For Multi-target Tracking.},
	Year = {2015}}

@article{denseflow,
	Author = {A. Wedel and T. Brox and T. Vaudrey and C. Rabe and U. Franke and D. Cremers},
	Date-Added = {2012-11-13 00:15:22 +0000},
	Date-Modified = {2012-11-13 00:16:22 +0000},
	Journal = {IJCV},
	Title = {Stereoscopic scene flow computation for 3D motion understanding},
	Year = {2011}}

@article{wojekcvpr2011,
	Author = {C. Wojek and S. Walk and S. Roth and B. Schiele},
	Date-Added = {2012-11-07 19:51:45 +0000},
	Date-Modified = {2012-11-15 02:56:10 +0000},
	Journal = {CVPR},
	Title = {Monocular 3D Scene Understanding with Explicit Occlusion Reasoning},
	Year = {2011}}

@article{Wolpert92,
	Author = {David H. Wolpert},
	Journal = {Neural Networks},
	Pages = {241--259},
	Title = {Stacked Generalization},
	Volume = {5},
	Year = {1992}}

@article{wuijcv2007,
	Author = {B. Wu and R. Nevatia},
	Date-Added = {2010-12-01 17:06:17 +0100},
	Date-Modified = {2010-12-01 17:07:07 +0100},
	Journal = {IJCV},
	Number = {2},
	Owner = {laurale},
	Timestamp = {2015.04.16},
	Title = {Detection and tracking of multiple, partially occluded humans by bayesian combination of edgelet part detectors},
	Volume = {75},
	Year = {2007}}

@article{wuwacv2009,
	Author = {Z. Wu and N.I. Hristov and T.H. Kunz and M. Betke},
	Date-Added = {2011-10-31 19:55:40 +0100},
	Date-Modified = {2011-10-31 19:58:14 +0100},
	Journal = {WACV},
	Owner = {laurale},
	Timestamp = {2015.04.16},
	Title = {Tracking-Reconstruction or Reconstruction-Tracking? Comparison of two multiple hypothesis tracking approaches to interpret 3D object motion from several camera views},
	Year = {2009}}

@article{wuiccv2009,
	Author = {Z. Wu and N. I. Hristov and T. L. Hedrick and T. H. Kunz and M. Betke},
	Journal = {ICCV},
	Owner = {laurale},
	Timestamp = {2015.02.05},
	Title = {Tracking a Large Number of Objects from Multiple Views},
	Year = {2009}}

@article{wucvpr2011,
	Author = {Z. Wu and T.H. Kunz and M. Betke},
	Date-Added = {2011-10-31 19:57:16 +0100},
	Date-Modified = {2011-10-31 19:58:42 +0100},
	Journal = {CVPR},
	Owner = {laurale},
	Timestamp = {2015.04.16},
	Title = {Efficient track linking methods for track graphs using network-flow and set-cover techniques},
	Year = {2011}}

@article{xiangiccv2015,
	Author = {Y. Xiang and A. Alahi and S. Savarese},
	Date-Added = {2015-11-04 13:59:53 +0000},
	Date-Modified = {2015-11-04 14:00:24 +0000},
	Journal = {ICCV},
	Title = {Learning to Track: Online Multi-Object Tracking by Decision Making},
	Year = {2015}}

@article{xueccv2012,
	Author = {C. Xu and C. Xiong and J.J. Corso},
	Journal = {ECCV},
	Owner = {laurale},
	Timestamp = {2015.04.16},
	Title = {Streaming hierarchical video segmentation},
	Year = {2012}}

@article{yamaguchicvpr2011,
	Author = {K. Yamaguchi and A.C. Berg and L.E Ortiz and T.L. Berg},
	Date-Added = {2011-07-06 20:29:18 +0200},
	Date-Modified = {2011-07-06 20:29:53 +0200},
	Journal = {CVPR},
	Owner = {laurale},
	Timestamp = {2015.04.16},
	Title = {Who are you with and where are you going?},
	Year = {2011}}

@article{yangcvpr2012,
	Author = {B. Yang and R. Nevatia},
	Date-Added = {2013-10-22 14:58:50 +0000},
	Date-Modified = {2013-10-22 14:59:49 +0000},
	Journal = {CVPR},
	Title = {An online learned CRF model for multi-target tracking},
	Year = {2012}}

@article{yangiccv2007,
	Author = {M. Yang and T. Yu and Y. Wu},
	Date-Added = {2010-12-01 15:24:03 +0100},
	Date-Modified = {2010-12-01 15:24:43 +0100},
	Journal = {ICCV},
	Owner = {laurale},
	Timestamp = {2015.04.16},
	Title = {Game-theoretic multiple target tracking},
	Year = {2007}}

@article{yeffeticcv2009,
	Author = {L. Yeffet and L. Wolf},
	Date-Added = {2013-10-27 15:58:48 +0000},
	Date-Modified = {2013-10-27 15:59:17 +0000},
	Journal = {ICCV},
	Title = {Local trinary patterns for human action recognition},
	Year = {2009}}

@article{yoonwacv2015,
	Author = {J. Yoon and H. Yang and J. Lim and K. Yoon},
	Journal = {IEEE Winter Conference on Applications of Computer Vision (WACV)},
	Owner = {laurale},
	Timestamp = {2015.04.07},
	Title = {Bayesian Multi-Object Tracking Using Motion Context from Multiple Objects},
	Year = {2015}}

@article{yucvpr2007,
	Author = {Q. Yu and G. Medioni and I. Cohen},
	Date-Added = {2011-11-21 16:21:56 +0100},
	Date-Modified = {2011-11-21 16:22:38 +0100},
	Journal = {CVPR},
	Owner = {laurale},
	Timestamp = {2015.04.16},
	Title = {Multiple target tracking using spatio-temporal {M}arkov chain {M}onte {C}arlo data association},
	Year = {2007}}

@article{zagoruyko2015,
	Author = {S. Zagoruyko and N. Komodakis},
	Journal = {CVPR},
	Title = {Learning to compare image patches via convolutional Neural Networks},
	Year = {2015}}

@article{zamireccv2012,
	Author = {A.R. Zamir and A. Dehghan and M. Shah},
	Journal = {ECCV},
	Owner = {laurale},
	Timestamp = {2015.10.28},
	Title = {GMCP-Tracker: Global Multi-object Tracking using Generalized Minimum Clique Graphs},
	Year = {2012}}

@article{Zbontar2015,
	Author = {J. Zbontar and Y. LeCun},
	Journal = {CVPR},
	Title = {Computing the Stereo Matching Cost With a Convolutional Neural Network},
	Year = {2015}}

@article{zhangcvpr2008,
	Author = {L. Zhang and Y. Li and R. Nevatia},
	Date-Added = {2010-10-21 17:47:16 +0200},
	Date-Modified = {2010-10-21 18:01:25 +0200},
	Journal = {CVPR},
	Owner = {laurale},
	Timestamp = {2015.04.16},
	Title = {Global data association for multi-object tracking using network flows},
	Year = {2008}}

@misc{bmtt2015,
	Journal = {WACV},
	Owner = {laurale},
	Timestamp = {2015.04.16},
	Title = {1st Workshop on Benchmarking Multi-Target Tracking},
	Url = {http://www.igp.ethz.ch/photogrammetry/bmtt2015/home.html},
	Year = {http://www.igp.ethz.ch/photogrammetry/bmtt2015/home.html},
	Bdsk-Url-1 = {http://www.igp.ethz.ch/photogrammetry/bmtt2015/home.html}}

@misc{gurobi,
	Date-Added = {2011-11-20 13:37:09 +0100},
	Date-Modified = {2011-11-20 13:42:08 +0100},
	Owner = {laurale},
	Timestamp = {2015.02.25},
	Title = {Gurobi Library},
	Url = {www.gurobi.com},
	Urldate = {http://sourceforge.net/apps/wordpress/dwsolver/},
	Year = {www.gurobi.com},
	Bdsk-Url-1 = {www.gurobi.com}}

@misc{rmrc2014,
	Owner = {laurale},
	Timestamp = {2015.02.13},
	Title = {Reconstruction Meets Recognition Challenge},
	Url = {http://ttic.uchicago.edu/~rurtasun/rmrc/index.php},
	Year = {http://ttic.uchicago.edu/~rurtasun/rmrc/index.php},
	Bdsk-Url-1 = {http://ttic.uchicago.edu/~rurtasun/rmrc/index.php}}


\begin{thebibliography}{10}
\providecommand{\url}[1]{#1}
\csname url@samestyle\endcsname
\providecommand{\newblock}{\relax}
\providecommand{\bibinfo}[2]{#2}
\providecommand{\BIBentrySTDinterwordspacing}{\spaceskip=0pt\relax}
\providecommand{\BIBentryALTinterwordstretchfactor}{4}
\providecommand{\BIBentryALTinterwordspacing}{\spaceskip=\fontdimen2\font plus
\BIBentryALTinterwordstretchfactor\fontdimen3\font minus
  \fontdimen4\font\relax}
\providecommand{\BIBforeignlanguage}[2]{{%
\expandafter\ifx\csname l@#1\endcsname\relax
\typeout{** WARNING: IEEEtran.bst: No hyphenation pattern has been}%
\typeout{** loaded for the language `#1'. Using the pattern for}%
\typeout{** the default language instead.}%
\else
\language=\csname l@#1\endcsname
\fi
#2}}
\providecommand{\BIBdecl}{\relax}
\BIBdecl

\bibitem{lealcvpr2014}
L.~Leal-Taix{\'e}, M.~Fenzi, A.~Kuznetsova, B.~Rosenhahn, and S.~Savarese,
  ``Lerning an image-based motion context for multiple people tracking,''
  \emph{CVPR}, 2014.

\bibitem{milantpami2014}
A.~Milan and S.~R.~R. Schindler, ``Continuous energy minimization for
  multitarget tracking,'' \emph{TPAMI}, 2014.

\bibitem{zamireccv2012}
A.~Zamir, A.~Dehghan, and M.~Shah, ``Gmcp-tracker: Global multi-object tracking
  using generalized minimum clique graphs,'' \emph{ECCV}, 2012.

\bibitem{dollartpami2014}
P.~Dol{\'l}ar, R.~Appel, S.~Belongie, and P.~Perona, ``Fast feature pyramids
  for object detection,'' \emph{TPAMI}, 2014.

\bibitem{sadeghieccv2014}
M.~Sadeghi and D.~Forsyth, ``30hz object detection with dpm v5,'' \emph{ECCV},
  2014.

\bibitem{rennips2015}
S.~Ren, R.~G. K.~He, and J.~Sun, ``Faster r-cnn: Towards real-time object
  detection with region proposal networks,'' \emph{Neural Information
  Processing Systems (NIPS)}, 2015.

\bibitem{milancvpr2015}
A.~Milan, L.~Leal-Taix{\'e}, K.~Schindler, and I.~Reid, ``Joint tracking and
  segmentation of multiple targets,'' in \emph{CVPR}, 2015.

\bibitem{fragkiadakieccv2012}
K.~Fragkiadaki, W.~Zhang, G.~Zhng, and J.~Shi, ``Two-granularity tracking:
  mediating trajectory and detections graphs for tracking under occlusions,''
  \emph{ECCV}, 2012.

\bibitem{chencvpr2014}
S.~Chen, A.~Fern, and S.~Todorovic, ``Multi-object tracking via constrained
  sequential labeling,'' \emph{CVPR}, 2014.

\bibitem{broxeccv2010}
T.~Brox and J.~Malik, ``Object segmentation by long term analysis of point
  trajectories,'' \emph{ECCV}, 2010.

\bibitem{xueccv2012}
C.~Xu, C.~Xiong, and J.~Corso, ``Streaming hierarchical video segmentation,''
  \emph{ECCV}, 2012.

\bibitem{galltpami2011}
J.~Gall, A.~Yao, N.~Razavi, L.~van Gool, and V.~Lempitsky, ``Hough forests for
  object detection, tracking and action recognition,'' \emph{TPAMI}, 2011.

\bibitem{partbased}
P.~Felzenszwalb, R.~Girshick, D.~McAllester, and D.~Ramanan, ``Object detection
  with discriminatively trained part based models,'' \emph{TPAMI}, 2010.

\bibitem{khantpami2005}
Z.~Khan, T.~Balch, and F.~Dellaert, ``Mcmc-based particle filtering for
  tracking a variable number of interacting targets,'' \emph{TPAMI}, 2005.

\bibitem{berclazcvpr2006}
J.~Berclaz, F.~Fleuret, and P.~Fua, ``Robust people tracking with global
  trajectory optimization,'' \emph{CVPR}, 2006.

\bibitem{jiangcvpr2007}
H.~Jiang, S.~Fels, and J.~Little, ``A linear programming approach for multiple
  object tracking,'' \emph{CVPR}, 2007.

\bibitem{zhangcvpr2008}
L.~Zhang, Y.~Li, and R.~Nevatia, ``Global data association for multi-object
  tracking using network flows,'' \emph{CVPR}, 2008.

\bibitem{pirsiavashcvpr2011}
H.~Pirsiavash, D.~Ramanan, and C.~Fowlkes, ``Globally-optimal greedy algorithms
  for tracking a variable number of objects,'' \emph{CVPR}, 2011.

\bibitem{berclaztpami2011}
J.~Berclaz, F.~Fleuret, E.~T{\"u}retken, and P.~Fua, ``Multiple object tracking
  using k-shortest paths optimization,'' \emph{TPAMI}, 2011.

\bibitem{dehghancvpr2015}
A.~Dehghan, S.~Assari, and M.~Shah, ``Gmmcp-tracker: Globally optimal
  generalized maximum multi clique problem for multiple object tracking,''
  \emph{CVPR}, 2015.

\bibitem{brendelcvpr2011}
W.~Brendel, M.~Amer, and S.~Todorovic, ``Multiobject tracking as maximum weight
  indepen- dent set,'' \emph{CVPR}, 2011.

\bibitem{lealcvpr2012}
L.~Leal-Taix{\'e}, G.~Pons-Moll, and B.~Rosenhahn, ``Branch-and-price global
  optimization for multi-view multi-object tracking,'' \emph{CVPR}, 2012.

\bibitem{buttcvpr2013}
A.~Butt and R.~Collins, ``Multi-target tracking by {L}agrangian relaxation to
  min-cost network flow,'' \emph{CVPR}, 2013.

\bibitem{leibeiccv2007}
B.~Leibe, K.~Schindler, and L.~van Gool, ``Coupled detection and trajectory
  estimation from multi-object tracking,'' \emph{ICCV}, 2007.

\bibitem{izadiniaeccv2012}
H.~Izadinia, I.~Saleemi, W.~Li, and M.~Shah, ``(mp)2t: Multiple people multiple
  parts tracker,'' \emph{ECCV}, 2011.

\bibitem{wojekcvpr2011}
C.~Wojek, S.~Walk, S.~Roth, and B.~Schiele, ``Monocular 3d scene understanding
  with explicit occlusion reasoning,'' \emph{CVPR}, 2011.

\bibitem{tangiccv2013}
S.~Tang, M.~Andriluka, A.~Milan, K.~Schindler, S.~Roth, and B.~Schiele,
  ``Learning people detectors for tracking in crowded scenes,'' \emph{ICCV},
  2013.

\bibitem{chari2015pairwise}
V.~Chari, S.~Lacoste-Julien, I.~Laptev, and J.~Sivic, ``On pairwise costs for
  network flow multi-object tracking,'' in \emph{Proceedings of the IEEE
  Conference on Computer Vision and Pattern Recognition}, 2015, pp. 5537--5545.

\bibitem{Seguin16}
G.~Seguin, P.~Bojanowski, R.~Lajugie, and I.~Laptev, ``Instance-level video
  segmentation from object tracks,'' in \emph{Proc. CVPR}, 2016.

\bibitem{benfoldcvpr2011}
B.~Benfold and I.~Reid, ``Stable multi-target tracking in real-time
  surveillance video,'' \emph{CVPR}, 2011.

\bibitem{tomasiklt1991}
C.~Tomasi and T.~Kanade, ``Detection and tracking of point features,''
  CMU-CS-91-132, Tech. Rep., 1991.

\bibitem{fragkiadakicvpr2011}
K.~Fragkiadaki and J.~Shi, ``Detection free tracking: Exploiting motion and
  topology for segmenting and tracking under entanglement,'' \emph{CVPR}, 2011.

\bibitem{choiiccv2015}
W.~Choi, ``Near-online multi-target tracking with aggregated local flow
  descriptor,'' \emph{ICCV}, 2015.

\bibitem{torresanieccv2008}
L.~Torresani, V.~Kolmogorov, and C.~Rother, ``Feature correspondence via graph
  matching: models and global optimization,'' \emph{ECCV}, 2008.

\bibitem{charicvpr2015}
V.~Chari, S.~Lacoste-Julien, I.~Laptev, and J.~Sivic, ``On pairwise costs for
  network flow multi-object tracking,'' \emph{CVPR}, 2015.

\bibitem{bansal2004correlation}
N.~Bansal, A.~Blum, and S.~Chawla, ``Correlation clustering,'' \emph{Machine
  Learning}, vol.~56, no. 1-3, pp. 89--113, 2004.

\bibitem{bagon2011large}
S.~Bagon and M.~Galun, ``Large scale correlation clustering optimization,''
  \emph{arXiv preprint arXiv:1112.2903}, 2011.

\bibitem{shi2000normalized}
J.~Shi and J.~Malik, ``Normalized cuts and image segmentation,'' \emph{TPAMI},
  vol.~22, no.~8, pp. 888--905, 2000.

\bibitem{cour2004normalized}
T.~Cour, S.~Yu, and J.~Shi, ``Normalized cut segmentation code. copyright 2004
  university of pennsylvania,'' \emph{Computer and Information Science
  Department}, 2004.

\bibitem{li2007noise}
Z.~Li, J.~Liu, S.~Chen, and X.~Tang, ``Noise robust spectral clustering,'' in
  \emph{ICCV}.\hskip 1em plus 0.5em minus 0.4em\relax IEEE, 2007, pp. 1--8.

\bibitem{zelnik2004self}
L.~Zelnik-Manor and P.~Perona, ``Self-tuning spectral clustering,'' in
  \emph{Advances in neural information processing systems}, 2004, pp.
  1601--1608.

\bibitem{motchallenge:arxiv:2015}
L.~Leal-Taix{\'e}, A.~Milan, I.~Reid, S.~Roth, and K.~Schindler, ``Motchallenge
  2015: Towards a benchmark for multi-target tracking,''
  \emph{arXiv:1504.01942}, 2015.

\bibitem{clear}
R.~Kasturi, D.~Goldgof, P.~Soundararajan, V.~Manohar, J.~Garofolo, M.~Boonstra,
  V.~Korzhova, and J.~Zhang, ``Framework for performance evaluation for face,
  text and vehicle detection and tracking in video: data, metrics, and
  protocol,'' \emph{TPAMI}, vol.~31, no.~2, 2009.

\bibitem{licvpr2009}
Y.~Li, C.~Huang, and R.~Nevatia, ``Learning to associate: hybrid boosted
  multi-target tracker for crowded scene,'' \emph{CVPR}, 2009.

\bibitem{dollarpami2014}
P.~Doll\'ar, R.~Appel, S.~Belongie, and P.~Perona, ``Fast feature pyramids for
  object detection,'' \emph{PAMI}, 2014.

\bibitem{pellegriniiccv2009}
S.~Pellegrini, A.~Ess, K.~Schindler, and L.~van Gool, ``You'll never walk
  alone: modeling social behavior for multi-target tracking,'' \emph{ICCV},
  2009.

\bibitem{yamaguchicvpr2011}
K.~Yamaguchi, A.~Berg, L.~Ortiz, and T.~Berg, ``Who are you with and where are
  you going?'' \emph{CVPR}, 2011.

\bibitem{lealiccv2011}
L.~Leal-Taix{\'e}, G.~Pons-Moll, and B.~Rosenhahn, ``Everybody needs somebody:
  Modeling social and grouping behavior on a linear programming multiple people
  tracker,'' \emph{ICCV. 1st Workshop on Modeling, Simulation and Visual
  Analysis of Large Crowds}, 2011.

\bibitem{yang2016temporal}
M.~Yang and Y.~Jia, ``Temporal dynamic appearance modeling for online
  multi-person tracking,'' \emph{Computer Vision and Image Understanding},
  2016.

\bibitem{kimiccv2015}
C.~Kim, F.~Li, A.~Ciptadi, and J.~Rehg, ``Multiple hypothesis tracking
  revisited: Blending in modern appearance model,'' \emph{ICCV}, 2015.

\bibitem{xiangiccv2015}
Y.~Xiang, A.~Alahi, and S.~Savarese, ``Learning to track: Online multi-object
  tracking by decision making,'' \emph{ICCV}, 2015.

\bibitem{wangbmvc2015}
S.~Wang and C.~Fowlkes, ``Learning optimal parameters for multi-target
  tracking.'' \emph{BMVC}, 2015.

\bibitem{mclaughlinwacv2015}
N.~McLaughlin, J.~M.~D. Rincon, and P.~Miller, ``Enhancing linear programming
  with motion modeling for multi-target tracking.'' \emph{WACV}, 2015.

\bibitem{rezatofighiiccv2015}
H.~Rezatofighi, A.~Milan, Z.~Zhang, Q.~Shi, and I.~R. A.~Dick, ``Joint
  probabilistic data association revisited,'' \emph{ICCV}, 2015.

\bibitem{milantpami2016}
A.~Milan, K.~Schindler, and S.~Roth, ``Multi-target tracking by
  discrete-continuous energy minimization,'' \emph{TPAMI}, 2016.

\bibitem{yoonwacv2015}
J.~Yoon, H.~Yang, J.~Lim, and K.~Yoon, ``Bayesian multi-object tracking using
  motion context from multiple objects,'' \emph{IEEE Winter Conference on
  Applications of Computer Vision (WACV)}, 2015.

\bibitem{dicleiccv2013}
C.~Dicle, O.~Camps, and M.~Sznaier, ``The way they move: Tracking targets with
  similar appearance,'' \emph{ICCV}, 2013.

\bibitem{bewleyicra2016}
A.~Bewley, L.~Ott, F.~Ramos, and B.~Upcroft, ``Alextrac: Affinity learning by
  exploring temporal reinforcement within association chains,'' \emph{ICRA},
  2016.

\bibitem{geigerpami2014}
A.~Geiger, M.~Lauer, C.~Wojek, C.~Stiller, and R.~Urtasun, ``3d traffic scene
  understanding from movable platforms,'' \emph{TPAMI}, 2014.

\bibitem{baecvpr2014}
S.~Bae and K.~Yoon, ``Robust online multi-object tracking based on tracklet
  confidence and online discriminative appearance learning,'' \emph{CVPR},
  2014.

\bibitem{soleraiccv2015}
F.~Solera, S.~Calderara, and R.~Cucchiara, ``Learning to divide and conquer for
  online multi-target tracking,'' \emph{ICCV}, 2015.

\end{thebibliography}

\pagebreak
\clearpage
\newpage
\appendix

In this appendix we detail the computation of the affinities between detections and dense point trajectories. \\

Note that we threshold very low affinities in order to speed the convergence of the clustering algorithm.

\vspace{0.5cm}

\noindent{\bf A1. Affinities between dense point tracks}\\
\label{sec:aff1}

The affinity matrix $ \mathbf{W}_\text{PP}$ is constructed considering motion and spatial proximity of dense point tracks.
In particular, let $T_{i},T_{j}$ be two dense point tracks. We define two affinities regarding the spatial distance and two affinities regarding their motion. Let $\mathcal{F}(i,j)$ denote the set of common frames between $T_{i}$ and $T_{j}$. We consider two cases: $\mathcal{F}(i,j) = \emptyset$ and $\mathcal{F}(i,j) \neq \emptyset$.
Now, let $\mathcal{F}(i,j) = \emptyset$. The optimal linking of dense point tracks in time has already been computed by the LP in Eq. \eqref{eq:LP}. During clustering we want to especially enforce coupling between different feature categories, hence
we set $W_\text{PP}(i,j)=1$ if there is a detection track connecting $T_{i}$ with $T_{j}$. Otherwise we set its value to $0$.

Now consider the case that $\mathcal{F}(i,j) \neq \emptyset$. We compute the affinity using 4 different aspects that we compare:

\begin{figure}[htp]
\centering
{\includegraphics[width=1\linewidth]{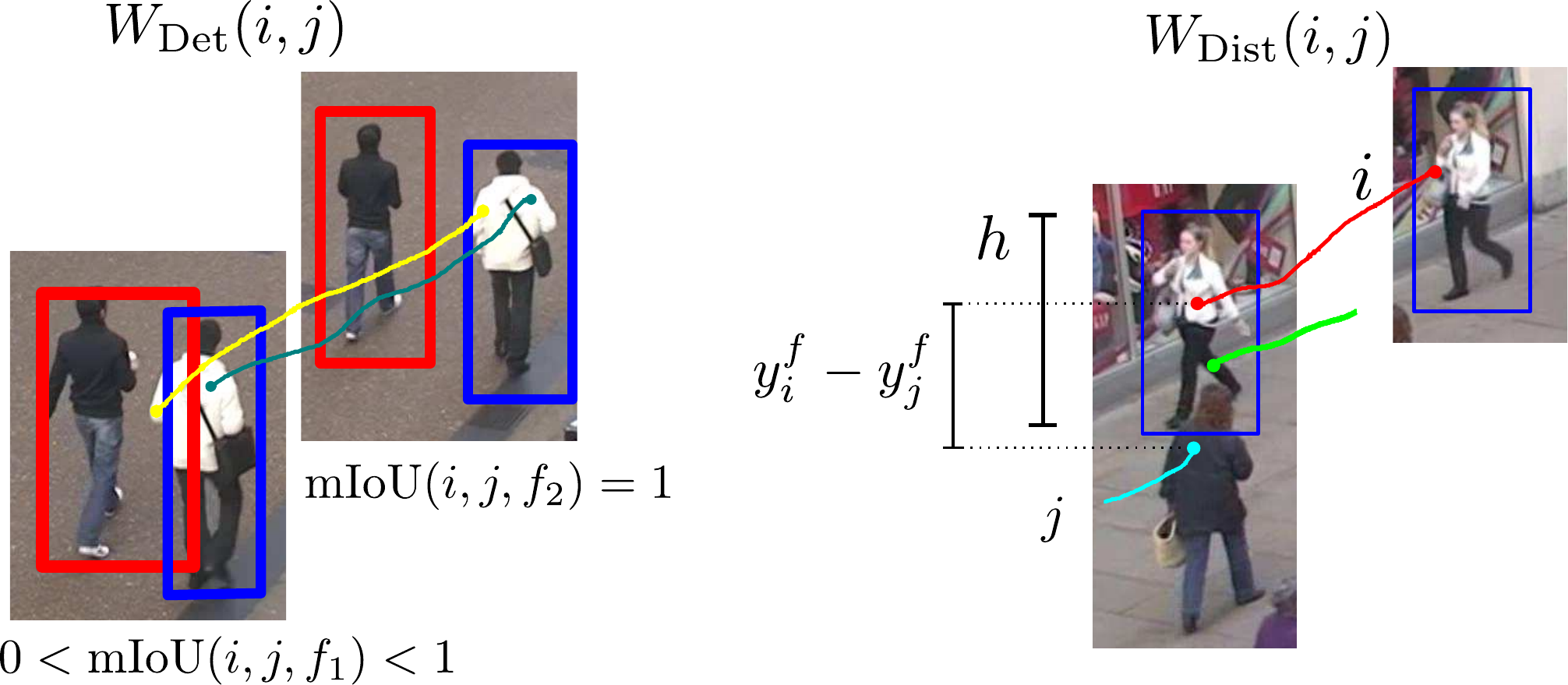}}
\caption{{\it Left}: Illustration of the $\mathbf{W}_\text{Det}$ definition. {\it Right}: the $\mathbf{W}_\text{dist}$ definition.}
\label{fig:det_dist}
\end{figure}

\begin{itemize}
\item The {\it detection} affinity matrix 
\begin{align}
W_\text{Det}(i,j) = \frac{\text{mIoU}(i,j)}{|\mathcal{F}(i,j)|}
\label{eq:Det}
\end{align}
measures how similar are the detections crossed by the dense point tracks, where 
\begin{align}
 \text{mIoU}(i,j)=\sum_{f \in \mathcal{F}(i,j)} \text{mIoU}(i,j,f),
\end{align}
is defined as the maximal intersection over union (IoU) between any two detections that intersect $T_{i}$ and $T_{j}$. If $T_{i}$ or $T_{j}$ do not have a detection at time stamp $f$, we set $\text{mIoU}(i,j,f) = 0.5$, as there is no scale available from which we can judge proximity. The left part of Figure \ref{fig:det_dist} explains the idea: the green and yellow dense points are close to each other but not in the same box, yet the affinity of being clustered together gets a non-zero value.\\
  
  \item The {\it distance} affinity matrix $\mathbf{W}_\text{Dist}$ measures directly the spatial distance of two dense tracks and weights it by the scale given by the detections. We equally weight information from width and height, hence we set $D(i,j) := 0.5 (D^H(i,j)+D^W(i,j))$. Then 
\begin{align}
W_\text{Dist}(i,j) = \begin{cases} D(i,j)& D^H(i,j),D^W(i,j) \geq 0.5 \\
0 & \text{otherwise}
\end{cases}
\label{eq:Dist}
\end{align}

Thereby $D^H$ measures the spatial proximity with respect to the height and $D^W$ with respect to the width, accordingly.
More specifically, for the height we define
\begin{align}
D^H(i,j):= \med\{D^H_{j}(i),D^H_{i}(j) \}.
\end{align}
Thereby, $D^H_{i}(j)$ judges whether $T_{j}$ should be clustered with $T_{i}$ on the basis of the detection height observed by $T_{i}$. 
Let $y_{i}^f$ be the y-position of $T_{i}$ at time $f$. We set  
\begin{align}
\hat{D}^H_{i}(j)= \med\left\{\frac{y_{i}^f - y_{j}^f}{\med_H(i)} \middle | f \in \mathcal{F}(i,j) \right\},
\end{align}
where $\med_{H}(i)$ is the median height of all boxes that intersect $T_{i}$. Now 
\begin{align}
D^H_{i}(j)= 
\begin{cases} 
1 & \text{if } \hat{D}^H_{i}(j) \leq \mu_{\text{Dist}} \\
\frac{\mathcal{N}\left(\hat{D}^H_{i}(j),\mu_{\text{Dist}},\sigma_\text{Dist}\right)}{\mathcal{N}\left(\mu_{\text{Dist}},\mu_{\text{Dist}},\sigma_\text{Dist}\right)} & \text{otherwise} \end{cases},
\end{align}
where $\mathcal{N}\left(x,\mu_{\text{Dist}},\sigma_{\text{Dist}}\right)$ denotes the normal probability density function evaluated at $x$ with mean value $\mu_{\text{Dist}}$ and variance $\sigma_{\text{Dist}}$.

The width affinities are defined accordingly, by replacing the $y$-coordinate with the $x$-coordinate and the median height by the median width.\\

\item Dense point tracks belonging to a person should have the same speed. Thus we define a {\it speed} affinity matrix 
\begin{align}
 W_\text{Speed}(i, j)  = \med \left\{ \frac{\min(v_i^f,v_j^f)}{\max(v_i^f,v_j^f)}   \right\},  
 \label{eq:Speed}
\end{align}
where  $f \in \mathcal{F}(i,j)$, and $v_i^f$ denotes the magnitude of the 2D velocity of $T_{i}$ at time stamp $f$.\\

\item Apart from the same speed, the dense point tracks should follow a similar direction. Hence, we compare angles between the velocity vectors of the tracks. We define the {\it angular} affinity as 
\begin{align}
W_\text{Angle}(i,j)=\frac{\mathcal{N}\left(\sphericalangle(i,j),0,\sigma_\text{angle}\right)}{\mathcal{N}\left(0,0,\sigma_\text{angle}\right)} , 
 \label{eq:Angle}
\end{align}
where $\sphericalangle(i,j)$ denotes the median angle between the velocity vectors of $T_{i}$ and $T_{j}$ at each common time stamp.

Finally, we combine speed and angle affinities with equal weight to create  the velocity $\mathbf{W}_\text{Velocity}$.
Since 2D velocities can be noisy, we reduce the weight of this affinity by transforming it linearly to the interval $[0.5,1]$.
\end{itemize}

Having defined these affinities we finally set
\begin{align}
W_\text{PP}(i,j)&=W_\text{Velocity}(i,j)\times \frac{1}{2}(W_\text{Dist}(i,j)+W_\text{Det}(i,j)).
\end{align}\\

\noindent{\bf A2. Affinities between a dense point track and a detection}\\

Given a dense point track ${T}_{i}$ and a detection $d_j$ of a detection track $T_j$, we compare them in two ways. First, we check for spatial intersection:
If $\mathcal{F}(i,j) \neq \emptyset$, we set
\begin{align}
W_\text{PD-intersect}(i,j)=\begin{cases}
1 & \text{if }  \mathbf{p}_i^f \text{lies in detection } d_j   \\
0 & \text{otherwise}.
\end{cases}
\end{align}
For the case $\mathcal{F}(i,j) = \emptyset$, we set the affinity to 0.5 for the reasons discussed above.
Furthermore, we compute $W_\text{PD-Link}(i,j)$ using the same terms as described for $\mathbf{W}_\text{PP}$. Finally, we combine the two terms:
\begin{align}
\mathbf{W}_\text{PD} = \dfrac{1}{2}(\mathbf{W}_\text{PD-intersect}+\mathbf{W}_\text{PD-Link}).
\end{align} \\

\noindent{\bf A3. Affinities between detections} \\

Comparison between detections is driven by the detection tracks as well as intersecting dense point tracks. Let $\mathcal{T}(d_i)$ be the set of dense point tracks intersecting $d_i$ and 
\[
r(i,j)=\frac{|\mathcal{T}(d_i) \cap \mathcal{T}(d_j)|}{|\mathcal{T}(d_i)|}.
\]
Then
\begin{align}
W_\text{DD-t}(i,j)=0.5*\begin{cases}
 r(i,j)+r(j,i) & |\mathcal{T}(d_i)|,|\mathcal{T}(d_j)| > 0  \\
1 & \text{otherwise}.
\end{cases}
\end{align}
A second term compares the whole span of the tracks:
\begin{align}
W_\text{DD-l}(i,j)=\begin{cases}
1 & \text{if } d_i, d_j \text{ are in the same track}   \\
0 & \text{otherwise}.
\end{cases}
\end{align}
Finally, we set $\mathbf{W}_\text{DD} = \mathbf{W}_\text{DD-t}  \mathbf{W}_\text{DD-l}$. \\

\end{document}